\documentclass[lettersize,journal]{IEEEtran}
\usepackage{amssymb}
\usepackage{amsmath}
\usepackage{graphicx}
\usepackage[colorlinks]{hyperref}
\hypersetup{
  citecolor = {blue}
}
\usepackage{siunitx}
\usepackage{pdflscape}
\usepackage{pdfpages}

\usepackage{enumitem}
\usepackage[normalem]{ulem}
\useunder{\uline}{\ul}{}
\usepackage{setspace} % \doublespacing
\usepackage[utf8]{inputenc}
\usepackage{rotating}
\usepackage{nicefrac}
\usepackage{xspace}
\usepackage{float}
\usepackage{caption}
\usepackage{subcaption}
\usepackage{rotating, lscape, longtable, booktabs, siunitx, multirow}
\usepackage[capitalise, noabbrev]{cleveref}
\usepackage{verbatim}
\usepackage{mathtools}
\usepackage{algorithm}% http://ctan.org/pkg/algorithm
\usepackage{algpseudocode}% http://ctan.org/pkg/algorithmicx
\usepackage{tkz-graph}
\usepackage{pifont}
\newcommand{\xmark}{\ding{55}}%

\usetikzlibrary{arrows,arrows.meta,shapes,decorations.pathmorphing, decorations.markings, positioning}
\let\svtikzpicture\tikzpicture
\def\tikzpicture{\noindent\svtikzpicture}
\newcommand{\uset}[1]{\ifmmode\left\{\,#1\,\right\}\else\{\,#1\,\}\fi}
\newcommand{\ulst}[1]{\ifmmode\left[\,#1\,\right]\else[\,#1\,]\fi}
\newcommand{\upar}[1]{\ifmmode\left(\,#1\,\right)\else(\,#1\,)\fi}
\newcommand{\uioc}[1]{\ifmmode\left(\,#1\,\right]\else(\,#1\,]\fi}
\newcommand{\uico}[1]{\ifmmode\left[\,#1\,\right)\else[\,#1\,)\fi}

\newcommand{\Vast}{\bBigg@{5}}

\begin{document}

\title{Semi-Supervised Health Index Monitoring with Feature Generation and Fusion}

\author{Ga\"{e}tan Frusque${}^{1}$, Ismail Nejjar${}^{1}$, Majid Nabavi${}^{2}$ and Olga Fink${}^{1}$ 
        % <-this % stops a space
\thanks{This study was supported by the Swiss Innovation Agency (lnnosuisse) under grant number: 47231.1 IP-ENG.}% <-this % stops a space
\thanks{G. Frusque, I. Nejjar and O. Fink are with Laboratory of Intelligent Maintenance and Operations Systems, EPFL, Lausanne, gaetan.frusque@epfl.ch, ismail.nejjar@epfl.ch olga.fink@epfl.ch. Majid Nabavi is with Oerlikon METCO, Wohlen, Majid.Nabavi@oerlikon.com.
}
\thanks{Manuscript sent to Transactions on Reliability.}}

\maketitle

\begin{abstract}
The Health Index (HI) is crucial for evaluating system health, important for tasks like anomaly detection and Remaining Useful Life (RUL) prediction of safety-critical systems. {\color{black}Real-time, meticulous monitoring of system conditions is essential, especially in manufacturing high-quality and safety-critical components, like spray coating}. However, acquiring accurate health status information (HI labels) in real scenarios can be difficult or costly because it requires continuous, precise measurements that fully capture the system's health. As a result, using datasets from systems run-to-failure, which provide limited HI labels at just the healthy and end-of-life phases, becomes a practical approach. {\color{black}
We employ the the Deep Semi-supervised Anomaly Detection (DeepSAD) embeddings  to tackle the challenge of extracting features associated with the system’s health state}
Additionally, we introduce a diversity loss to further enrich the DeepSAD embeddings.  We also propose applying an alternating projection algorithm with isotonic constraints to transform the embedding into a normalized HI with an increasing trend. Validation on the PHME2010 milling dataset, a recognized benchmark with ground truth HIs confirms the efficacy of our proposed HIs estimations.
Our methodology is further  applied to monitor wear states of thermal spray coatings using high-frequency voltage. Our contributions facilitate  more accessible and reliable HI estimation, particularly in scenarios  where obtaining ground truth HI labels is impossible.
\end{abstract}


\section{Introduction}

Thermal spray coating is a materials processing technique used to apply protective or decorative coatings to various surfaces. Major customers for thermal spray products include manufacturers of jet engines, gas turbines, and automobiles. {\color{black}Due to the criticality of these applications, substantial resources are invested by these customers in evaluating coating quality and addressing issues associated with  parts that have  low-quality coatings. Common problems include uneven material thickness across the surface, rough surface finish, poor adhesion, and high porosity.}  These defects are often indicative of the operational state and overall health condition of the thermal spray coating system. The Health Index (HI), alternatively referred to as a health indicator, serves as a critical tool for assessing these  conditions \cite{fink2020potential}, \cite{lei2018machinery}. It frequently serves as a important metric for subsequent prognostics and health management (PHM) tasks, such as anomaly detection \cite{michau2017deep}, condition monitoring \cite{hsu2023comparison}, and prediction of remaining useful life \cite{wang2021remaining}. 
{\color{black}Estimating the HI of a thermal spray system  using  model-based approaches requires substantial modelling efforts and is often impractical  in real-worlds applications. Data-driven approaches help alleviate the burden of modeling.} 

{\color{black}The data-driven strategies for estimating the HI of an engineered system can be categorized into three main groups: supervised, unsupervised, or semi-supervised. Supervised HI estimation requires either direct or indirect measurements of one or several damage indicators that align with the most commonly worn component of the system. For instance, in the case of the PHME 2010 Milling dataset \cite{li2009fuzzy}, which includes run-to-failure data from a cutting tool, the degradation of each flank wear on the three cutting edges is measured using a microscope after each cutting pass. Since  it is a controlled experiment focused on flank degradation, it is assumed that  no other factors are impacting the system's health in this experimental setup. Therefore, the HI of the system can be approximated using these measured damage indicators. Using the damage indicators as labels, } numerous regression models, such as stacked sparse autoencoder \cite{he2022milling}, informer encoder \cite{li2022intelligent}, Wiener process \cite{liu2022three}, and bi-directional LSTM \cite{zhou2022milling}, have been employed to predict the HI of the milling system. However, datasets with ground truth measurements of the HI, as showcased in this example, are rare because obtaining these labels is often prohibitively costly for companies, or there may be no direct way to measure the health condition.

Unsupervised HI estimation is a more commonly employed approach. It involves learning solely from a dataset assumed to represent a healthy state. By understanding the distribution of the healthy state, it is possible to infer a HI in real-time by assessing how much the current measurements deviate from this healthy distribution \cite{rombach2022contrastive}. This approach is primarily utilized for anomaly detection, and one of the most frequently applied methods here is One-Class Classifiers (OCC), often in combination with deep learning AE architectures. Examples of  models for HI estimation include Autoencoders \cite{hsu2023comparison, jin2022condition}, Support Vector Data Description \cite{chao2023health, frusque2023non} or OCC with Extreme Machine Learning \cite{michau2017deep}.
However, translating the OCC's output into a meaningful HI measure can be challenging, as it is sensitive to variations in the system's wear and  operating conditions. Additionally, we do not leverage potential information about the data gathered when we have doubts about its current health status or during failure, which could be valuable in enhancing the final HI estimation.

{\color{black}Semi-supervised learning methods have been primarily used in the context of RUL prediction\cite{che2022semi}  and anomaly detection \cite{zhao2014graph}. These methods are particularly valuable  for HI estimation, as they often  leverage expert knowledge from the entire lifecycle,  utilizing  run-to-failure data.
This may involve integrating physical simulations of the system's healthy operation \cite{sun2023adaptive} or incorporating regularization to enforce expected behavior in the HI \cite{moradi2023intelligent}.
Some studies focus on semi-supervised HI estimation based on multi-objective optimization.
By employing multiple objective functions, this approach captures how well HI adheres to specific properties such as high trendability, monotonicity and prognosability\cite{moradi2023intelligent}, as well as robustness \cite{chao2023health, chen2022deep}. 
One way to achieve  a monotone HI is to add an isotonic constraint, which ensures that the index only increases or decreases. This property can either be used as a post-processing operation to ensure the final index has monotonic evolution \cite{wang2021remaining} or as an additional constraint added to the loss function \cite{nieves2022semi}.
However, multi-objective optimization often requires fine-tuning numerous hyperparameters and can be challenging due to the complexity of the loss function.

Other semi-supervised approaches use deep neural networks to obtain an embedding that encapsulates information about the health state of the system by learning from data representing both the healthy and faulty states of the system. In  one study \cite{jang2023deep}, a stacked denoising autoencoder is used to construct a HI based on user-defined monitoring conditions. In another study \cite{chen2023interpretable}, a regularized autoencoder approach with latent space variance maximization is employed to encode the health condition into features that characterize the transition between healthy and degraded data. The Deep Semi-supervised Anomaly Detection model (DeepSAD) \cite{ruff2019deep, delise2023deep} was designed to create embeddings that differentiate between
healthy and abnormal classes, demonstrating effectiveness in anomaly detection tasks. {\color{black} Our proposed approach can be generally applied with any semi-supervised learning methods. We selected DeepSAD because it offers the advantages of providing end-to-end representations, demonstrating strong performance against other semi-supervised approaches \cite{han2022adbench}, and featuring a loss function with a limited number of hyperparameters.
 }}

{\color{black}Currently, there is a lack of suitable tools for monitoring the HI of thermal spray coatings continuously. While collecting a run-to-failure dataset from the thermal spray coating system is feasible, acquiring labels on the health state, which can only indirectly be inferred based on the coating quality, is prohibitively expensive. In this work, we  utilize information pertaining solely to extreme health conditions — namely, healthy and worn-out states — and train the model to infer the health condition between these two extremes. DeepSAD semi-supervised learning algorithm is particularly relevant in our context since we consider the specific case where the knowledge is an approximate estimate of when a degraded state has been reached and when a system is in its healthy state, typically at the time it was first put into operation. Consequently, only a binary damage indicator is provided for the task and we then assume that, given the labels for those two binary states along with the full degradation trajectory that is not labeled, we aim to propose an algorithm that will be able to interpolate between these extremes and predict the entire health evolution.}

{\color{black}As our first contribution, we extend the application of DeepSAD into the domain of HI estimation.} Instead of directly using the norm of the DeeepSAD model output as a HI, we propose considering the embedding generated by DeepSAD as a condition indicator that needs to be integrated to construct the HI.
The limitations of using the norm of the embedding as a HI are twofold. Firstly, interpreting the DeepSAD output as a HI can be challenging, similar to the OCC. Secondly, the norm output often remains very low during healthy periods, hindering the capture of variations in the wear state during these phases. This limitation can impede  the practical utility of the HI for tasks such as RUL prediction or anomaly detection.
{\color{black}However, the embedding generated by the DeepSAD model can be low-rank, characterized by redundancy and dimensions composed only of zero values.} To diversify the condition indicators derived from the DeepSAD embedding, we propose incorporating  a diversity loss.

As our second contribution, we introduce a novel approach to HI estimation through feature fusion, employing an isotonic alternating projection algorithm. This concept involves projecting an index into both the input feature subspace and the space representing the ideal HI. We define the ideal HI as a collection of trajectories adhering to specific  properties: they must start at 0 and reach 1 when the system is considered worn out, exhibiting a monotonic increase.

In the first step, we evaluate the proposed methodology using  the PHME 2010 milling machine benchmark dataset \cite{li2009fuzzy}. Notably, the ground truth labels, the damage indicators that serve as a proxy for the real health condition, are never used during the training of our model. However, they play a crucial role in validating the performance of the generated HIs. We assess the quality of our estimated HIs by examining their correlation with the ground truth HIs. We investigate  whether the variations in HI values between different systems hold meaningful significance. {\color{black}Finally, we assess the quality of the obtained HI using standard criteria such as monotonicity, trendability, and prognosability \cite{moradi2023intelligent}.} In the second step, we apply our approach to a real-world dataset: a spray coating dataset collected by Oerlikon Metco. In this use case, we analyze the time-series voltage signal generated by the thermal spray gun to estimate the remaining useful lifetime of the gun's critical components. The quality of this index is evaluated by comparing it to domain expert indications and standard HI quality metrics.

\section{Method}
\subsection{HI generation using Embedding Diversified DeepSAD }
\subsubsection{DeepSAD}

We consider a training dataset denoted as $\mathbf{X}=\{\mathbf{x}_1,...,\mathbf{x}_{N_l}, \mathbf{x}_{N_l+1}, \mathbf{x}_{N} \}$, where there is a total of $N = N_l + N_u$ samples. Each sample comprises feature vectors $\mathbf{x}$ in $\mathbb{R}^{F}$ of dimension $F$. Here, $N_l$ represents the number of labeled samples, and $N_u$ represents the number of unlabeled samples. The labels are denoted by $l \in \{1,-1\}$, with a value of 1 assigned for samples that are a realization of a healthy system, and a value of -1 assigned when the sample represents a realization of a system with a severe fault. Samples in-between are then unlabelled.

The Deep Semi-supervised Anomaly Detection method \cite{ruff2019deep} aims to discover a transformation $\phi_\theta$ using a neural network with weights $\theta$ to effectively separate healthy and unlabeled samples from the abnormal ones. The primary objective of the  of the DeepSAD method is to minimize the volume of a hypersphere centered at $\alpha$,  encompassing healthy samples, while ensuring that abnormal samples lie outside this hypersphere. We denote the DeepSAD loss function as $\mathcal{L}_{\rm DS}(\mathbf{X}; \theta)$, and the parameters $\theta$ are determined by minimizing this loss function. It can be expressed as follows:
\begin{align}
    \underset{\theta,\alpha  }{ \rm argmin}\hspace{0.5cm} &   \sum_{i=1}^{N_l}   \mid \mid \phi_\theta(\mathbf{x_i}) - \mathbf{\alpha}   \mid \mid_F^{2l_i} \\ \nonumber & \hspace{-1cm} + \mu \sum_{j=N_l+1}^{N} \mid \mid \phi_\theta(\mathbf{x_j}) - \mathbf{\alpha} \mid \mid_F^{2} + \nu \mid \mid \theta \mid \mid_F^2  
\end{align}

The parameter $\mu$ serves as a hyperparameter that determines the extent to which unlabeled samples are incorporated   within the hypersphere that encompasses healthy samples. In contrast, $\nu$ is a crucial hyperparameter that regularizes the neural network's weights,  preventing overfitting.

Finally, we represent the DeepSAD embedding of dimension $K$ for the sample $\mathbf{x}_i$ as $\mathbf{y}_i = \phi(\mathbf{x}_i) - \alpha$.

\subsubsection{Generating embedding with more diversity}
In practical scenarios, the DeepSAD embedding $\mathbf{Y}$ often exhibits a low rank structure, with repeated dimensions containing  identical information, and some dimensions remaining null. This behavior arises from the DeepSAD objective function, which primarily emphasizes the norm of its embeddings rather than their actual values. To address this issue, this work introduces an enrichment approach for the DeepSAD embedding by introducing a novel diversity loss function.

Referring to $\mathbf{C} = (\mathbf{Y}^T \mathbf{Y})$ as the Gram matrix of the DeepSAD embeddings, the suggested diversity regularization can be expressed as follows:
\begin{align}
 \mathcal{L}_{\rm Diversity}(\mathbf{C}) = - {\rm ln}({\rm det}( \mathbf{C} ) ) +  {\rm trace}(\mathbf{C} )
\end{align}
Here, ${\rm ln}({\rm det}(\bullet))$ represents the natural logarithm of the matrix determinant. The revised loss function, incorporating diversity regularization into the DeepSAD model, is named Diversity-DeepSAD and denoted as 2DS and can be expressed as follows:
\begin{align}\label{divDS}
    \underset{\theta  }{ \rm argmin}\hspace{0.5cm} & \mathcal{L}_{\rm DS}(\mathbf{X}; \theta) + \lambda  \mathcal{L}_{\rm Diversity}(\mathbf{C})
\end{align}
where $\lambda$ is a hyperparameter related to the diversity regularisation. 

\begin{figure}[h]
\centering
\includegraphics[width=0.4\textwidth]{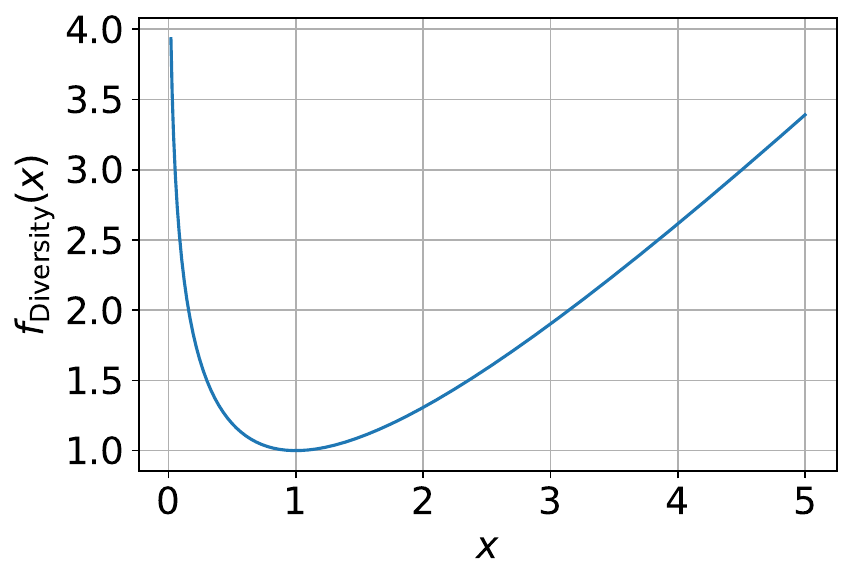}%width=0.7\textwidth
\caption{Diversity function apply to each eigenvalues of $\mathbf{C}$. }\label{fig:logdet}
\end{figure}

The rationale behind the proposed diversity regularization can be grounded in its frequent application in precision matrix estimation, often utilizing graphical loss algorithms. In this context, it resembles the task of estimating a precise precision matrix for an isotropic multivariate Gaussian distribution \cite{friedman2008sparse, frusque2020regularized}. The objective of the proposed diversity regularization is achieved when $\mathbf{C} = \mathbf{I}$, as demonstrated by observing that the gradient of $ \mathcal{L}_{\rm Diversity}$ with respect to the matrix $\mathbf{C}$ is as follows:
\begin{align}
\nabla \mathcal{L}_{\rm Diversity}(\mathbf{C}) = -  \mathbf{C}^{-1} + \mathbf{I}
\end{align}
Consequently,  enforcing the matrix $\mathbf{C}$ to approach the identity matrix implies that the various embeddings of the DeepSAD model should exhibit orthogonal and distinct behaviors.
Another perspective is to examine the eigenvalues of the proposed diversity regularization. Let $\sigma_i$ denote the $i^{\rm th}$ eigenvalue of the matrix $\textbf{C}$. The diversity regularization can then be expressed as follows:
\begin{align}
 \mathcal{L}_{\rm Diversity}(\mathbf{C}) = \sum_{i=1}^F  \sigma_i  -{\rm ln}( \sigma_i )  
\end{align}
This regularization entails applying the function $f(x) = x - \ln(x)$ to each eigenvalue, as depicted in Figure ~\ref{fig:logdet}. Notably, this function encourages the matrix $\mathbf{C}$ to maintain full rank, promoting diversity among trajectories while preventing eigenvalues from becoming  excessively high.

\subsection{Feature fusion using an Alternating Projection Algorithm with isotonic contraints}
\subsubsection{Proposed feature fusion methodology}
When considering a DeepSAD embedding, denoted as $\mathbf{Y}$, the objective is to determine the optimal combination of these features to construct a HI, denoted as $\mathbf{h}$.
For this section, the matrix $\mathbf{Y}$ has to be organized in a sequence corresponding to the order in which samples from the analyzed system were recorded. The time index is represented as $t \in {1, \ldots, T}$.

We propose constructing the HI using an alternating projection algorithm with the objective of finding a HI, denoted $\mathbf{h} \in \mathbb{R}^{T}$, that closely approximates the space of HIs we consider as ideal. This ideal HI space,  denoted as ${\rm E}{\rm I}$ and is defined as \\ $\{{{\rm E}_{\rm I} =  \mathbf{z} \mid z_t \leq 0 \text{ if } t \leq T_d, z_t \geq 1 \text{ if } t \geq T_f, z_{t+1} \geq z_t } \}$. In essence, it implies that an ideal HI should have values below 0 when $t$ is less than the time threshold $T_d$, representing periods when we assume our samples originate from a healthy system. Conversely, we anticipate the HI to have values above 1 when $t$ exceeds the time threshold $T_f$, signifying periods when we consider our samples to come from a degraded system. Furthermore, we expect the HI to exhibit a monotonically increasing trend, capturing changes related to wear rather than shifts in operating conditions. This constraint is referred to as isotonic regression, as introduced in works such as \cite{tang1991extension,lanza2011statistical}, and has recently been applied in \cite{wang2021remaining} for HI denoising. The optimization algorithm involves finding the regressor $\mathbf{w} \in \mathbb{R}^K$ such that:
\begin{align}\label{met:merge}
\underset{\mathbf{w},\mathbf{z}  }{ \rm argmin}\hspace{0.5cm} & \mid \mid \mathbf{h} - \mathbf{z}   \mid \mid_F^2 + \beta {\rm R}(\mathbf{w}) \\ 
\hspace{0.5cm} {\rm s.t.} \hspace{0.5cm} &\mathbf{h} = \mathbf{Y} \mathbf{w}   \nonumber \\
\hspace{0.5cm} & \mathbf{z} \in {\rm E}_{\rm I} \nonumber 
\end{align}

In this context, $\mathbf{z}$ represents a HI that falls within the set ${\rm E}_{\rm I}$, and $\mathbf{R}(\bullet)$, with a hyperparameter $\beta$, acts  as a potential regularization function designed to prevent overfitting. This regularization function can take the form of ridge regularization, denoted as $\mathbf{R}(\mathbf{w}) = ||\mathbf{w}||_F^2$, but it can also be extended to incorporate lasso or elastic net regularization if the feature space has high dimensionality, denoted as $K$.

\subsubsection{Algorithm}

To address the optimization problem presented in Equation~\ref{met:merge}, we propose an alternating approach \cite{escalante2011alternating}, in which we iteratively optimize the regressors $\mathbf{w}$ and the ideal HI $\mathbf{z}$. When optimizing $\mathbf{w}$ while keeping $\mathbf{z}$ fixed, the optimization problem in Equation~\ref{met:merge} transforms into the following:
\begin{align}\label{met:reg2}
\underset{\mathbf{w} }{ \rm argmin}\hspace{0.5cm} & \mid \mid  \mathbf{Y} \mathbf{w} - \mathbf{z}   \mid \mid_F^2 + \beta {\rm R}(\mathbf{w}) 
\end{align}
Depending on the type of regularization used, denoted as ${\rm R}(\mathbf{w})$, this process involves solving a ridge, lasso, or elastic net regression. Conversely, when optimizing $\mathbf{z}$ while keeping $\mathbf{w}$ fixed, the optimization problem in Equation~\ref{met:merge} transforms into: 
\begin{align}
\underset{\mathbf{z}  }{ \rm argmin}\hspace{0.5cm} & \mid \mid \mathbf{h} - \mathbf{z}   \mid \mid_F^2  \\ 
\hspace{0.5cm} & \mathbf{z} \in {\rm E}_{\rm I}   \nonumber
\end{align}
 
This step involves directly projecting the HI $\mathbf{h}$ onto the space of the ideal HI. To ensure  the HI's monotonic increase, we perform an isotonic regression, utilizing the Pool Adjacent Violator Algorithm \cite{wang2021remaining, tang1991extension} which is notably efficient with a complexity of O(t).

\begin{figure}[h]
\centering
\includegraphics[width=0.4\textwidth]{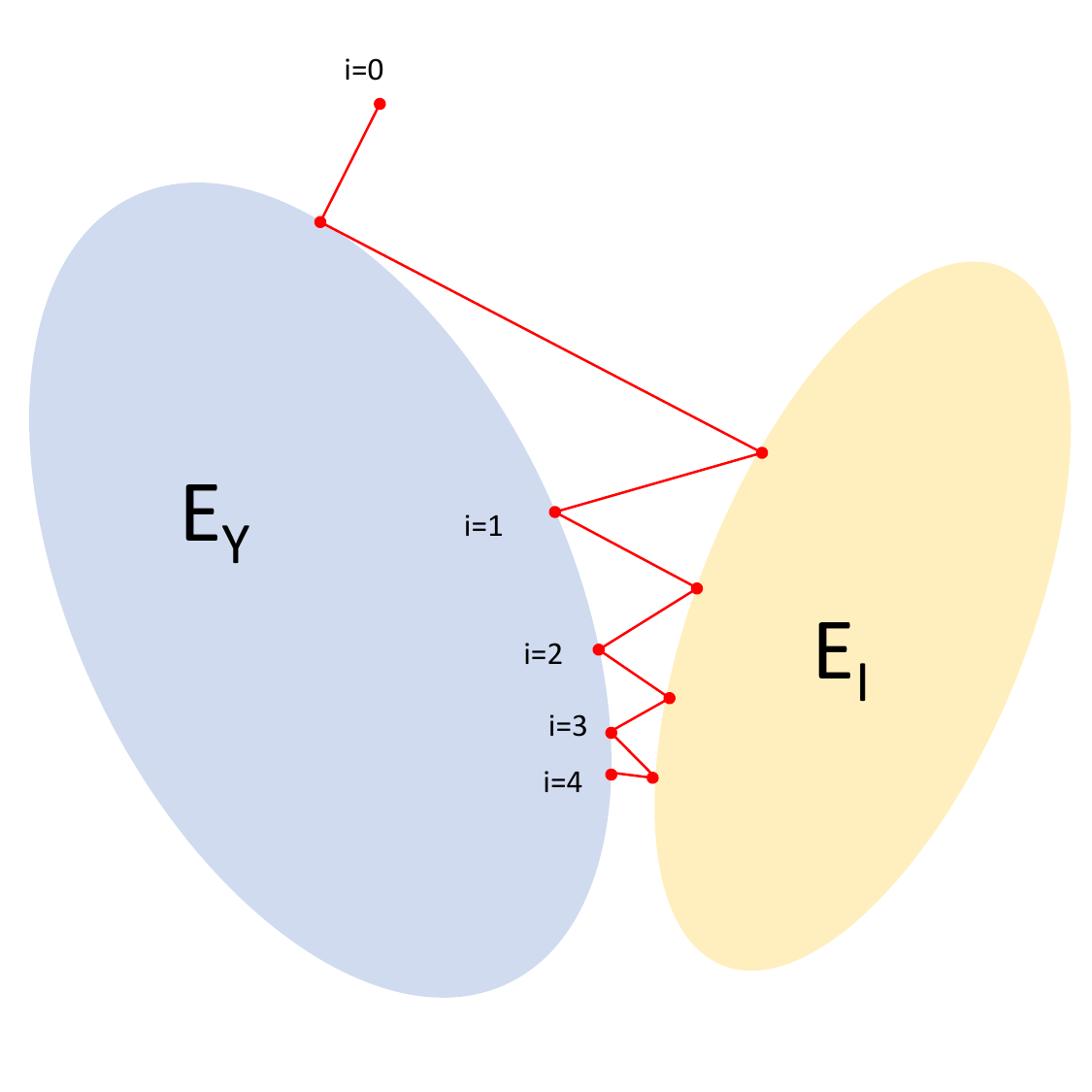}%width=0.5\textwidth
\caption{Schematic illustration of the alternating projection algorithm for a 2-D space. $E_I$ represents the space of perfect health indicators, while $E_Y$ denotes the space generated by the dataset. }\label{fig:AltProj}
\end{figure}

It is worth noting that when $\beta=0$, this process effectively projects the HI simultaneously onto the subspace defined by the features $\mathbf{Y}$ and the space of ideal HIs ${\rm E}_{\rm I}$, as illustrated in Figure~\ref{fig:AltProj}. However, since the subspace generated by $\mathbf{Y}$ may encompass ${\rm E}_{\rm I}$ in cases where the condition $K<<F$ is not met, this can potentially lead to less relevant solutions that are highly sensitive to the algorithm's initialization. In such scenarios, the regularization ${\rm R}(\mathbf{w})$ becomes particularly crucial.

The algorithm of the proposed Alternating Projection Algorithm with Isotonic Constraint (APAIC) is presented in Algorithm.~\ref{alg:cap}
\begin{algorithm}
\caption{Algorithm to solve the optimisation problem \ref{met:merge}}\label{alg:cap}
\begin{algorithmic}
\Require Dataset $\mathbf{Y}$ and hyperparameter $\beta$
    \State $\mathbf{w}= \mathbf{1}$
    \For{$i \in$   $\{1,...,I\}$}
        \State Solve the regression problem equation \ref{met:reg2}
        \State $\mathbf{h} =  \mathbf{Y} \mathbf{w} $
        \State $\mathbf{z}=\mathbf{h}$
        \State $z_t=0$ for $t \in \{1,...,T_d\}$ if $z_t>0$
        \State $z_t=1$ for $t \in \{T_f,...,T\}$ if $z_t<1$
        \State Perform isotonic projection on $\mathbf{z}$ using the PAVA algorithm \cite{wang2021remaining}
    \EndFor
\end{algorithmic}
\end{algorithm}
\subsubsection{Training and real-time HI construction}
 
In practice, our optimization algorithm combines data from both the training and validation datasets with our test dataset. This approach is necessary because it is not feasible for the test dataset to determine the degraded time threshold $T_f$, as we construct the HI specifically to estimate it. Therefore, when we denote $\mathbf{Y}^{(v)}$ as $v$ different validation or training datasets, and $\mathbf{Y}_{:t}$ as the first $t$ recorded samples of the investigated system, the optimization problem in Equation~\ref{met:merge} transforms into:
\begin{align}\label{eq:time}
\underset{\mathbf{w}, \{\mathbf{z}^{(1)},.., \mathbf{z}^{(K)}\}, \mathbf{z} }{ \rm argmin}\hspace{0.5cm} & \sum_{k=1}^{K} \mid \mid \mathbf{h}^{(k)} - \mathbf{z}^{(k)}   \mid \mid_F^2 + \mid \mid \mathbf{h}_{:t} - \mathbf{z}  \mid \mid_F^2 + \beta {\rm R}(\mathbf{w}) \\ 
\hspace{0.5cm} {\rm s.t.} \hspace{0.5cm} &\mathbf{h}^{(k)} = \mathbf{Y}^{(k)} \mathbf{w}  \nonumber \\
\hspace{0.5cm} & \mathbf{z}^{(k)} \in {\rm E}_{\rm I}  \nonumber \\
\hspace{0.5cm} &\mathbf{h}_{:t}= \mathbf{Y}_{:t} \mathbf{w}   \nonumber \\
\hspace{0.5cm}  & \mathbf{z} \in {\rm E}_{\rm I}^{\rm Test}   \nonumber
\end{align}

In this context, ${\rm E}_{\rm I}^{\rm Test} = {\{ \mathbf{z} \mid z_t \leq 0 \text{ if } t \leq T_d, z_{t+1} \geq z_t \} }$ represents the subset of ideal test HIs, excluding the worn-out condition. To address this optimization problem, Algorithm~\ref{alg:cap} can be applied. It involves concatenating features and ideal HIs for updating the regressor $\mathbf{w}$, while the projection onto the ideal subspace should be carried out separately for each system.

\section{Application on a benchmark dataset: The PHME 2010 milling wear datasets}
The International Prognostic and Health Management 2010 Challenge (PHM2010) Milling Wear Datasets address the issue of deterioration of milling tools and the continuous tracking of this wear within machining systems. In this section we propose to apply the proposed semi-supervised APAIC merging methodology on the 2DS features for HI estimation. The HI estimation is done here without using the labels provided by the dataset for the training of the models.
\subsection{Dataset}

\begin{figure}[h]
\centering
\begin{subfigure}[b]{0.45\textwidth}
        \centering \includegraphics[width=0.8\textwidth]{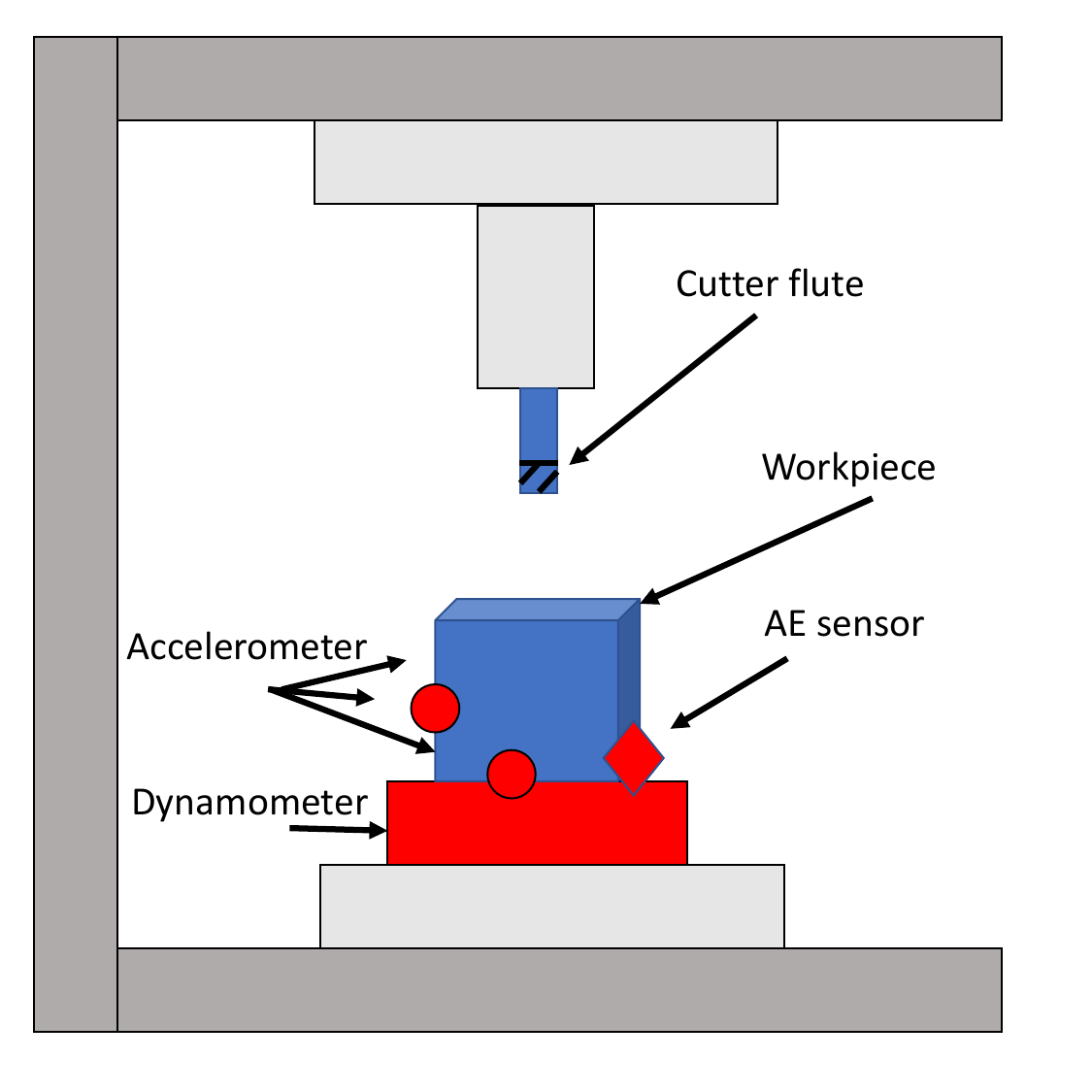}
        \caption{}
\end{subfigure}
\begin{subfigure}[b]{0.45\textwidth}
        \centering \includegraphics[width=\textwidth]{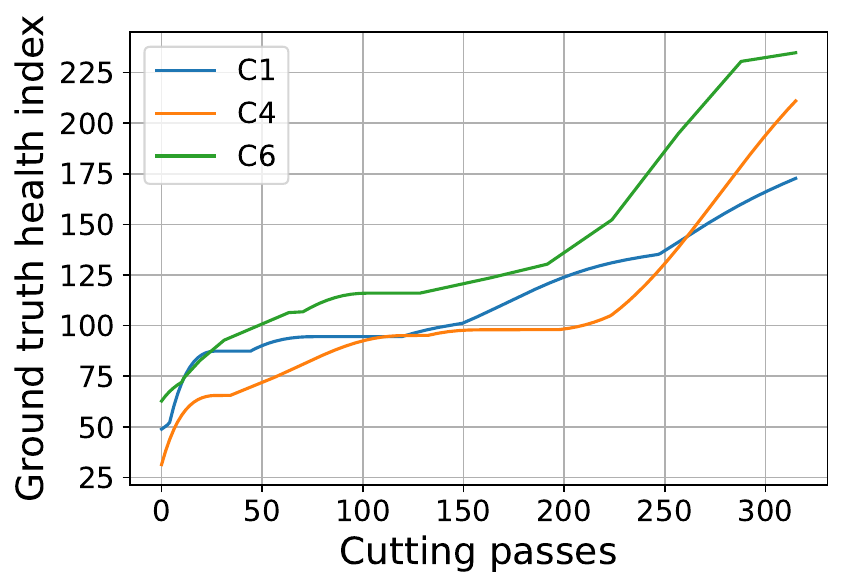}
        \caption{}
\end{subfigure}
    \caption{(a) PHME 2010 data acquisition experimental platform - (b) Ground truth HI for the three complete lifecycles dataset}\label{fig:Milling}
\end{figure}

The PHM2010 dataset originates from a high-speed computerized numerical control machine known as the Röders Tech RFM760. The dataset encompasses data collected from seven distinct sensors, measuring cutting forces, vibration, and acoustic emissions. Data acquisition for each channel occurred at a rate of 50 KHz. Figure~\ref{fig:Milling}(a) provides an illustration of the experimental data acquisition platform.

A dynamometer was installed between the machine table and the workpiece to measure cutting forces along three directions: x, y, and z. Additionally, three Kistler piezo accelerometers were positioned to monitor machine tool vibrations in the x, y, and z directions. Lastly, a Kistler Acoustic Emission sensor was employed to track high-frequency stress waves, and the data is provided as the root mean square of the acoustic emission. 

Following each cutting test, an offline measurement of the flank wear depth of the three individual flutes was conducted using a LEICA MZ12 microscope. The maximum wear depth observed serves as a valuable health state indicator for assessing the cutting tool.

In total, three milling experiments  with ground truth HIs were conducted (denoted as C1, C4, and C6). Figure~\ref{fig:Milling} (b) displays the full trajectories of these three HIs. 

{\color{black} For all methods, we utilize a cross-validation approach across the three datasets, where two datasets are used for training and the model is then tested on the remaining one.}
For additional information about the system, further details can be found in the references \cite{li2009fuzzy}, \cite{he2022milling}, and \cite{liu2022three}.

\subsection{Input features and metrics}

For each sensor modality, we transformed the raw sensor data into a mel spectrogram with 64 channels. We selected a window size of 0.1 seconds and a hop length of 0.1 seconds, this is the standard range of values used for Mel spectrogram computation \cite{purohit2019mimii}. The mel spectrograms from all sensor modalities were  then merged  along the feature dimension to form  the input feature vector $\mathbf{x}$ for each time step, resulting in a vector in dimension $F=448$.

\subsubsection{Metrics using the ground truth HI}
The goal of this study is to find a model that map the input feature $\mathbf{X}$ into an estimation $\hat{\mathbf{h}}$ of the ground truth HI obtained through microscopy. We focus on the average HI obtained for each cutting pass. To evaluate the quality of the estimated HI, we employ the following two metrics:
\begin{itemize}
    \item \textbf{Correlation}: 
    The correlation is important for evaluating the similarity between the shape of our estimated HI and that of the ground truth HI. The correlation score for any trajectory denoted as $c \in \{{\rm c1}, {\rm c4}, {\rm c6}\}$ is calculated as follows:
    \begin{equation}\label{EqCorr}
        {\rm Correlation} = \frac{\hat{\mathbf{h}}^t\mathbf{h}}{|| \hat{\mathbf{h}} ||_F || \mathbf{h} ||_F}
    \end{equation}
    \item \textbf{RMSE}: 
    The Root Mean Squared Error (RMSE) is used to assess  the relationship between the values of different HIs. In particular, if the ground truth HI value for one experiment exceeds  the values for another experiment, it should be reflected in the estimated HI. Since  our estimated HI values fall within the range of 0 to 1, we rescale them using the following operation:
    \begin{align}
        & \hat{\mathbf{h}'} = \hat{\mathbf{h}}M+m, \\
        & M=\frac{M_2-m_2}{M_1-m_1},\nonumber \\
        & m=m_2-m_1 M, \nonumber\\
        & m_1={\rm mean}([\hat{\mathbf{h}}^{\rm (c1)}_{100:150},\hat{\mathbf{h}}^{\rm (c4)}_{100:150},\hat{\mathbf{h}}^{\rm (c6)}_{100:150}]),\nonumber \\
        & m_2={\rm mean}([\mathbf{h^{\rm (c1)}}_{100:150},\mathbf{h^{\rm (c4)}}_{100:150},\mathbf{h^{\rm (c6)}}_{100:150}]), \nonumber \\
        & M_1={\rm max}([\hat{\mathbf{h}}^{\rm (c1)}, \hat{\mathbf{h}}^{\rm (c4)}, \hat{\mathbf{h}}^{\rm (c6)}]), \nonumber \\
        & M_2={\rm max}([\mathbf{h^{\rm (c1)}}, \mathbf{h^{\rm (c4)}}, \mathbf{h^{\rm (c6)}}]). \nonumber
    \end{align}
    Although it may appear complex, this equation essentially ensures  that both the ground truth and estimated HIs have identical means during the stationary period from 100 to 150, as well as matching maximum values across the three experiments. {\color{black}We choose values between 100 and 150 based on the ground truth index, ensuring that this range corresponds to a portion fully within the steady-state degradation period for these three indexes.} This operation simply entails applying the same affine transformation to the three estimated HIs, ensuring that their relative relationships remain unchanged. Consequently, for any experiment denoted as $c \in \{{\rm c1}, {\rm c4}, {\rm c6}\}$, the RMSE score can be expressed as follows:
    \begin{equation}
        {\rm RMSE} = || \hat{\mathbf{h}'} - \mathbf{h}||_F.
    \end{equation}
\end{itemize}

{\color{black}
\subsubsection{Standard unsupervised HI quality criteria}
As additional evaluation, we also consider the classical HI criteria \cite{chen2022deep} that are monotonicity, trendability and prognosability. 

$\bullet$ \textbf{Monotonicity} in data refers to the consistent upward or downward movement of a variable over time. Given that the wear of a system can only worsen over time, it is a critical property of a health indicator. Various metrics are utilized for assessing monotonicity. {\color{black}To account for minor variations due to noise and normalized scores between 0 and 1, we employ the modified Mann–Kendall monotonicity index given in \cite{eleftheroglou2020adaptive}.} The equation for the modified Mann-Kendall Monotonicity is as follows:}
\begin{equation}
\frac{\sum_{t=1}^{T} \sum_{\tau =t+1}^{T} \text{sign}(h_\tau - h_t) (\tau - t)}{\sum_{t=1}^{T} \sum_{\tau =t+1}^{T} (\tau - t)} 
\end{equation}
$\bullet$ {\color{black}\textbf{Trendability} refers to the characteristic of a dataset that allows for the identification of a similar trend between HIs of different units.} The trendability between the obtained HI from a model, considering the correlation Equation \ref{EqCorr}, can be expressed as \cite{moradi2023intelligent}:
\begin{equation}\label{trendability}
   \underset{k'  }{ \rm min}\hspace{0.5cm}  {\rm Corr}(\mathbf{h}^{(k)},\mathbf{h}^{(k')}) \hspace{0.5cm} k' \in \{ {\rm c1,c4,c6} \}
\end{equation}
$\bullet$ \textbf{Prognosability} indicates the variation in the HI at the failure point; its value distribution should significantly differ from that of the health state. Given that the flutes are in different health states at the end of the recording, we take that into consideration and evaluate the prognosability at time $t_P(k)$ such as \cite{moradi2023intelligent}:
\begin{equation}
    t_P(k) = \underset{t  }{ \rm argmin}\hspace{0.5cm} \mathbf{h}^{(k)}_t \geq \mathbf{h}^{(c1)}_T  .
\end{equation}
It corresponds to the moment when the HI first reaches the maximum value for the c1 milling machine, which exhibits the least wear. The prognosability index becomes: 
\begin{equation}\label{prog}
    {\rm exp}\left( \frac{\underset{k  }{{\rm std}}(h^{(k)}_{t_P(k)})}{\underset{k  }{{\rm mean}}(|h^{(k)}_{t_P(k)}-h^{(k)}_0|)} \right)  \hspace{0.5cm} k' \in \{ {\rm c1,c4,c6} \}
\end{equation}

\subsection{Performance of the APAIC merging algorithm}
\subsubsection{APAIC training}

Initially, we employ the APAIC algorithm directly on the raw features $\mathbf{X}$ without utilizing the DeepSAD algorithm for condition indicator estimation. For this analysis, we consider the average features for each cutting pass, totaling  $T=315$ cutting passes.

Subsequently, we proceed to directly determine the regressor $\mathbf{w}$ that satisfies Equation~\ref{eq:time} with $t=T=315$. For this purpose, we utilize two trajectories for the training dataset,  corresponding to the HIs projected into the space ${\rm E}_{\rm I}$ as described in Equation~\ref{eq:time}. Conversely, for the test experiment, the HIs are projected into the space ${\rm E}_{\rm I}^{\rm Test}$ since there is no available information regarding the end of life.

{\color{black} We set $T_d=50$ and $T_f=T-50=265$ to emulate a scenario where the expert's labeling to distinguish between healthy and worn-out parts is uncertain. As mentioned in \cite{liu2022three} and similarly to other mechanical systems, the milling machine undergoes three distinct wear periods characterized by different degradation dynamics: the incipient wear period, the steady-state period, and the deterioration period. The incipient wear period corresponds to the initial stage during which a cutting tool begins to experience gradual wear and degradation until the steady-state period. During the steady-state period, the wear dynamics are assumed to be linear and weak in comparison to the two other periods. Finally, the deterioration phase corresponds to the last period. In this phase, a higher degradation rate is initiated, which often concurs with the initiation of a defect or a fault, the progression under the high degradation rate continues until the cutting tool is completely deteriorated or is removed from the production line. Ideally, $T_d$ should encompass data from the first and the very beginning of the second period, while $T_f$ should never contain data from the first two periods, only from the degradation period, preferably when advanced degradation appears.}

{\color{black} In our approach to ridge regression, we empirically employed a ridge regularization with $\beta=0.05$. Given a substantial number of features, this parameter facilitates the model in conducting feature fusion that enhances its generalizability. Although a wide range of values can yield similar results, ridge regularization is important, as our model's performance diminishes when using a simple linear regression approach. The impact of this parameter was evaluated in appendix \ref{App:A1}.} In Figure~\ref{fig:Iso}, we provide an example illustrating the various updates to the HI when employing Algorithm~\ref{alg:cap}. The black line represents the initialization when we consider the sum of all features (we subtract the sum of all features after the first cutting pass from it, so it starts at 0).

During the initialization phase, the HI is not relevant, in contrast to the final iteration depicted by the red curve. In the end, we obtain a HI that is a monotonically increasing function remaining below 0 for the first 50 iterations and surpassing 1 for the last three iterations. The gradual convergence to the final solution for each iteration is indicated  by the color progression, ranging from dark blue to light green.
\begin{figure}[h]
\centering
\includegraphics[width=0.5\textwidth]{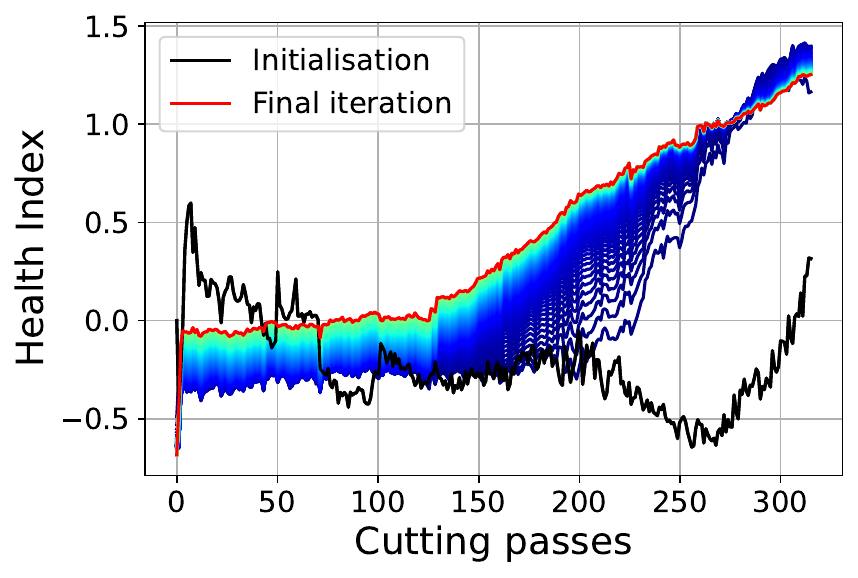}
    \caption{This visualization illustrates the evolution of the APAIC merging algorithm over 1000 iterations on a validation dataset. The algorithm's progression is depicted through a series of curves, transitioning from blue to green every 10 iterations. The initial state is represented in black, while the final result is highlighted in red.  }\label{fig:Iso}
\end{figure}

\subsubsection{Compared methodologies}
{\color{black}We conduct a comparison between our APAIC merging method and another approach that selects the best feature from the pool of 448 available features based on a specified criteria referred to as Multi-Objectif Feature "MO Feature".
In this context, feature selection is carried out with the objective of selecting the feature that maximizes the combined sum of the MK monotonicity, trendability, and prognosability indexes across the two training trajectory experiments.
Additionally, we compare our method to two oracle approaches, denoted as "Corr Oracle " and "RMSE Oracle" that select the feature directly according to the test dataset. In case of "Corr Oracle", our objective is to find the feature that minimizes the correlation score across all three trajectories simultaneously. Similarly, for "RMSE Oracle", the goal is to find the feature that minimizes the RMSE score across all three trajectories simultaneously. The Oracle methods are unrealistic in real-world scenarios and illustrates how the feature selection would perform if the data intended for the algorithm's application were available during training and could be utilized at that time.}

\subsubsection{Results}

{\color{black}The results for Correlation and RMSE are presented in Table~\ref{Table:Feature}. Notably, there is a significant disparity in RMSE and correlation scores between the "Corr Oracle" and "RMSE Oracle" approaches. This discrepancy underscores that distinct features are optimal for predicting the ground truth HI in each trajectory, and the same feature may exhibit varying behaviors and scales across different trajectories. As a consequence, the "MO Feature" feature exhibits the poorest performance in terms of RMSE because the features can behave differently between the training and test datasets.

Finally, we observe that employing the APAIC merging strategy results in both favorable Correlation scores and RMSE scores. We applying APAIC, we achieve the best RMSE score, improving from 27.5 to 24.3, which is better than the oracle "RMSE Oracle" selection strategy that uses the test dataset to find the best feature. This demonstrates that the APAIC strategy is the most reliable for HI estimation, maintaining  relevant relationships between features without access to the test dataset and HI ground truth labels.}
\begin{table*}
\center
\begin{tabular}{c|cccc|cccc}
  &  \multicolumn{4}{c|}{RMSE}  & \multicolumn{4}{c}{Correlation} \\
Method  & c1 & c4 & c6 & \textbf{Mean} & c1 & c4 & c6 & \textbf{Mean}\\
\hline
Corr Oracle &  28.88 &40.17 &65.05 &44.70 & 0.956 & 0.934 & 0.968 & \textbf{0.952} \\ 
RMSE Oracle &11.51 &41.42 &29.69 &27.54 & 0.975 & 0.883 & 0.821 & 0.893 \\ 
MO Feature & 21.93 &51.97 &67.45 &47.12  & 0.770 & 0.924 & 0.867 & 0.854 \\
APAIC Feature & 22.73 & 29.29 & 20.84 & \textbf{24.29} & 0.946 & 0.885 & 0.952 & 0.928 \\
\end{tabular} 
\caption{{\color{black}Correlation and adjusted Mean Squared Error (MSE) scores for the HI obtained using the APAIC merging method, the Oracle best raw features for all lifecycles based on Correlation "Corr Feature", the Oracle best raw feature for each lifecycle based on RMSE "RMSE Feature", and the best raw features for each lifecycle obtained based on HI criteria "MO Feature".}}\label{Table:Feature}
\end{table*}

\subsection{Performances of combining DeepSAD and APAIC}\label{MillingExp}
Based on the findings from the previous experiment, it is evident  that using raw features directly  for constructing the HI may have limitations. Therefore, our proposed approach first involves  using DeepSAD to directly create the HI. We then consider the embedding of DeepSAD as condition indicators related to the health state of the machine. These condition indicators are subsequently merged using the APAIC methodology. 
\subsubsection{DeepSAD training}
For DeepSAD training, the final 50 cutting passes from the training dataset are labeled  as abnormal samples (label -1 for DeepSAD), while the initial 50 samples are labeled as healthy labels (label 1 for DeepSAD). The remaining samples in the training dataset are considered unlabeled. Furthermore, for training DeepSAD, we include the initial 50 cutting passes from both the validation and test datasets as healthy samples with labels 1. {\color{black}The selection of the first and last 50 samples is comparable to the approach used for the APAIC algorithm, where $T_f=T_d=50$.}
We used the Adam optimizer with a training step size of $5e\text{-}4$ for 1000 epochs, utilizing a batch size of 128. {\color{black}The parameter $\nu$ was fixed at 0.1 to take into account that the unlabelled period may contains part that are already from the degradation phase. The weight decay parameter $\mu$ was initially set to 1 and empirically adjusted by gradually increasing it until consistent results were achieved during two simultaneous training sessions of the DeepSAD model with different initial seeds. The DeepSAD model's architecture comprises a three-layer dense neural network with 32 neurons each. We used the ReLU activation function for the first two layers and a linear activation function for the final layer. 
The embedding dimension is set to 16 to restrict the number of trajectories. The influence of this parameter is elaborated further in Appendix \ref{App:A2}.}
\subsubsection{Proposed approaches and comparison}
The APAIC merging strategy is applied to the embedding $\mathbf{Y}$ of the DeepSAD model. Unlike the straightforward utilization of Mel spectrogram features, the embedding $\mathbf{Y}$ encompasses multiple features that should be linked to the system's health status and can act as condition indicators. The resulting Health Indicator (HI) is derived by applying APAIC to these obtained condition indicators and is referred to as APAIC DeepSAD (ADS).
Additionally, we mimic the real-time HI estimation case, where incoming data is recorded on the fly. In this case, Equation~\ref{eq:time} is minimized several times for $t \in \{ \tau, 2\tau, ..., T \}$ with a step size of $\tau=30$. The final HI is obtained by concatenating the $\tau$ most recent steps of each computed HI. {\color{black}The choice of $\tau$ can be tailored to the specific application and the frequency required for monitoring the machine's health state.} This real-time variant of our proposed methodology is denoted as RADS.
Finally, we explore the scenario in which we train the 2DS model Equation.~\ref{divDS} using a fixed value of $\lambda=0.001$ to employ the diversity loss. 
{\color{black}$\lambda=0.001$ corresponds to having regularization within the same range of values as the DeepSAD loss function. The impact of this parameter is studied in appendix \ref{App:A3}. }This allows us to generate the HI using the 2DS, A2DS and RA2DS models. The various  terminologies for the proposed approaches are summarized in the Table~\ref{Table:Nomenc}

\begin{table}
\center
\begin{tabular}{c|ccc}
Method name & APAIC  & Real-Time  & Diversity \\
 & Eq.~\ref{eq:time} & Eq.~\ref{eq:time}  & loss Eq.~\ref{divDS} \\
\hline
DeepSAD &   \xmark &  \xmark &  \xmark  \\ 
ADS &  \checkmark &  \xmark &  \xmark  \\ 
RADS & \checkmark & \checkmark &  \xmark  \\
2DS &  \xmark &  \xmark& \checkmark  \\
A2DS &  \checkmark & \xmark & \checkmark  \\
RA2DS & \checkmark & \checkmark & \checkmark \\
\end{tabular} 
\caption{Terminology for the various methods}\label{Table:Nomenc}
\end{table}

For comparison with an unsupervised setting, we introduce a one-class classifier, the Support Vector Data Description (SVDD) \cite{tax2004support}. We consider the radial basis function kernel and empirically tune the hyperparameters $C$ and $\gamma$ based on the validation dataset, selecting the parameter combination that results in an estimated HI minimizing the distance from the ideal HI space ${\rm E}_{ \rm I}$. 

{\color{black}Furthermore, we compare our model against two semi-supervised approaches: a multi-objective optimization \cite{moradi2023intelligent} named MOO and an iterative pseudo-labeling procedure using isotonic constraints and binary damage indicator labels \cite{nieves2022semi} named "pseudo-label". Regarding MOO \cite{moradi2023intelligent}, we employ the same loss function, which is a weighted sum of MK monotonicity, trendability, and prognosability. 
However, the prognosability loss function differs from the prognosability score in Equation~\ref{prog}. As we do not have access to ground truth labels during training, we employed the standard prognosis loss function, where $t_P=T$ for all trajectories.
Since each criterion has a value range from 0 to 1, the weight is set to one for each criterion as proposed in the original paper \cite{moradi2023intelligent}. 
The neural network and training parameters used are  those of the DeepSAD model, with the exception that the last layer comprises only one neuron to produce the trajectory output. As for the "pseudo-label" procedure, the initialization of the pseudo labels mirrors that of the APAIC approach; label 0 represents the first 50 samples, while label 1 corresponds to the last 50 samples.}

\subsubsection{Results}
\begin{figure*}[htb!]
\centering
\begin{subfigure}[b]{0.42\textwidth}
        \centering \includegraphics[width=\textwidth]{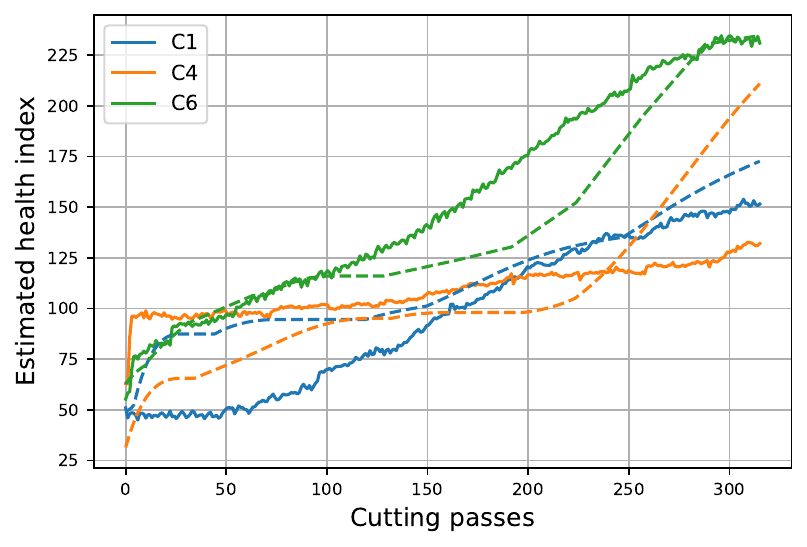}
        \caption{}
\end{subfigure}
\begin{subfigure}[b]{0.42\textwidth}
        \centering \includegraphics[width=\textwidth]{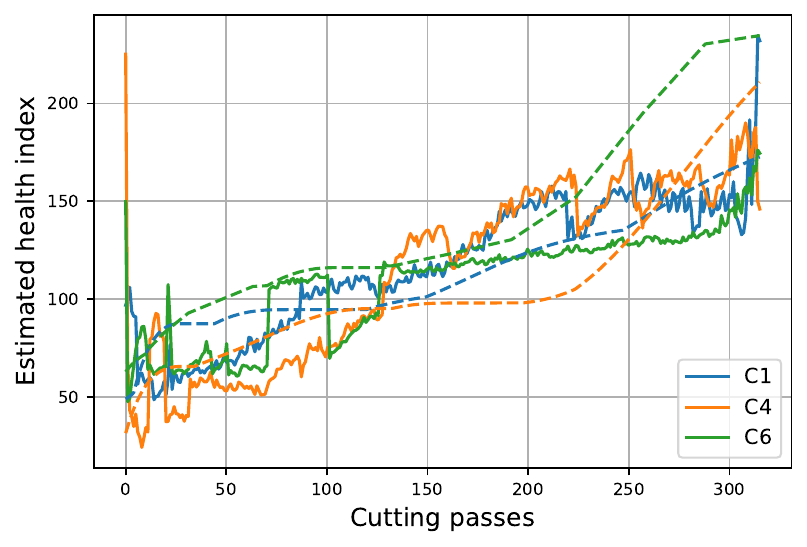}
        \caption{}
\end{subfigure}

\begin{subfigure}[b]{0.42\textwidth}
        \centering \includegraphics[width=\textwidth]{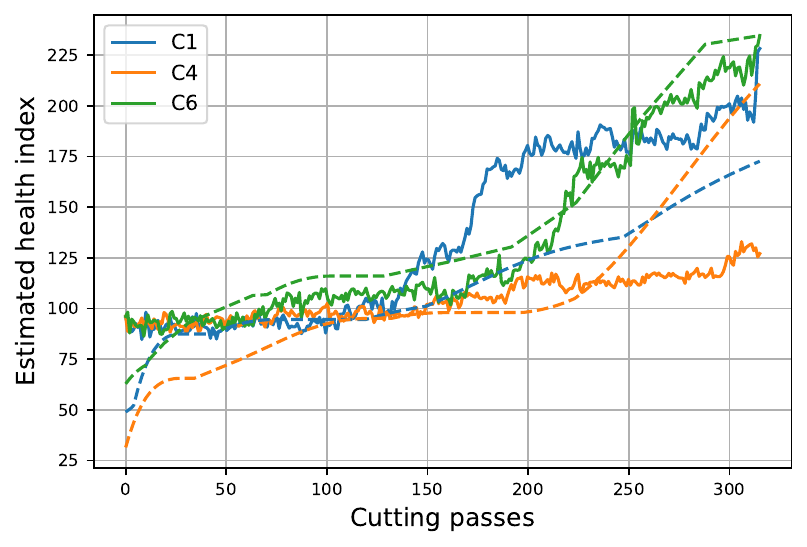}
        \caption{}
\end{subfigure}
\begin{subfigure}[b]{0.42\textwidth}
        \centering \includegraphics[width=\textwidth]{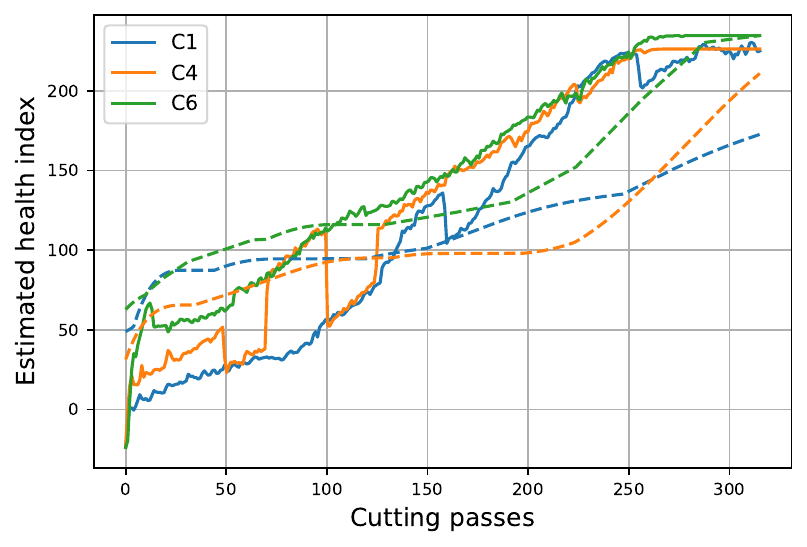}
        \caption{}
\end{subfigure}

\begin{subfigure}[b]{0.42\textwidth}
        \centering \includegraphics[width=\textwidth]{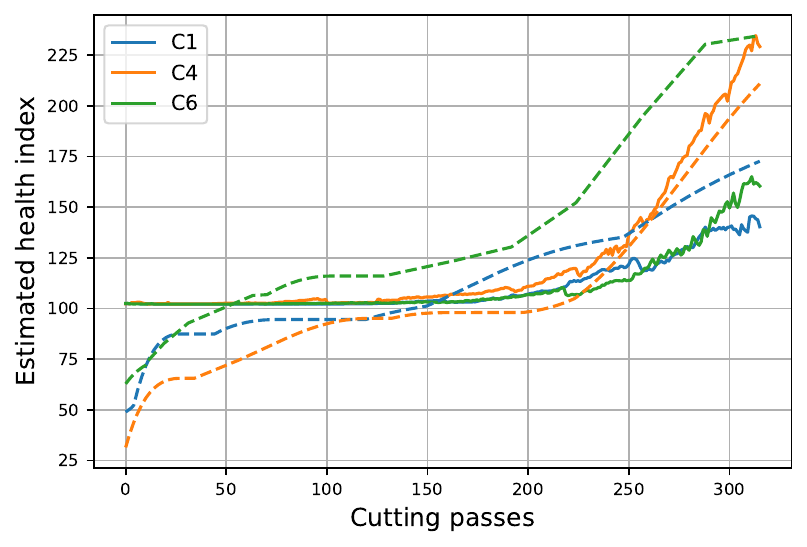}
        \caption{}
\end{subfigure}
\begin{subfigure}[b]{0.42\textwidth}
        \centering \includegraphics[width=\textwidth]{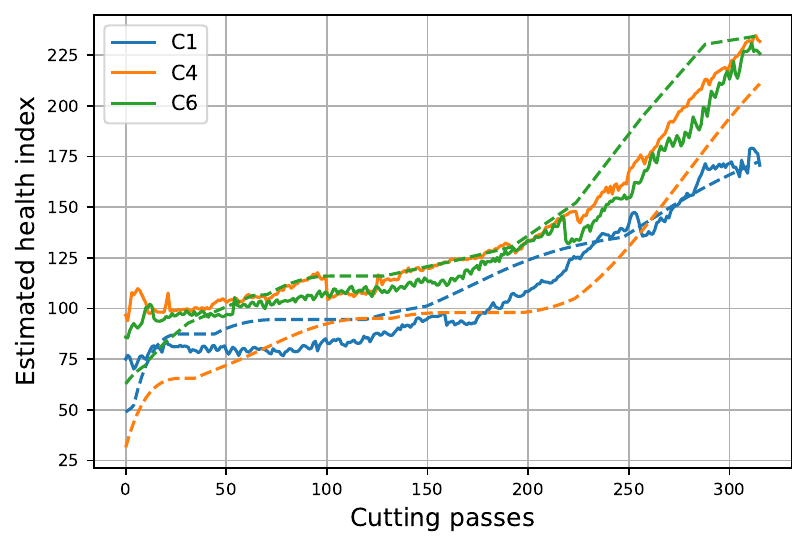}
        \caption{}
\end{subfigure}

\begin{subfigure}[b]{0.42\textwidth}
        \centering \includegraphics[width=\textwidth]{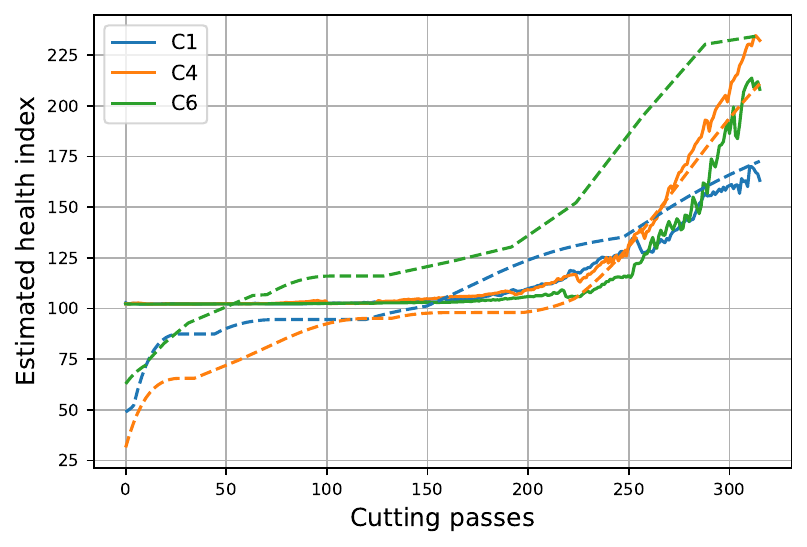}
        \caption{}
\end{subfigure}
\begin{subfigure}[b]{0.42\textwidth}
        \centering \includegraphics[width=\textwidth]{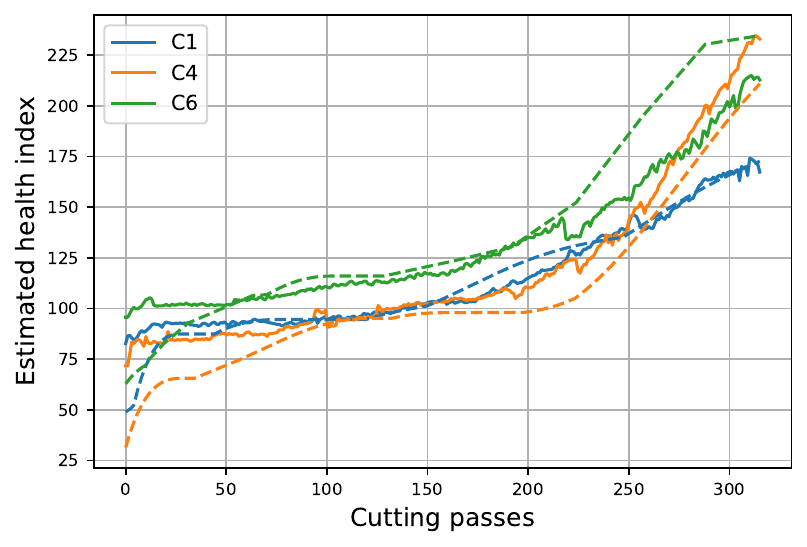}
        \caption{}
\end{subfigure}
    \caption{\color{black}Comparison between the estimated (solid line) and ground truth (dotted) for the following methods (a) APAIC Feature (b) SVDD \cite{tax2004support} (c) Pseudo-label \cite{nieves2022semi} (d) MOO \cite{moradi2023intelligent} (e) DeepSAD \cite{ruff2019deep} (f) RADS (g) 2DS (h) RA2DS}\label{fig:y}
\end{figure*}

{\color{black}Table~\ref{Table:Milling} displays the results comparing all methods. For the deep learning model, we present the mean and standard deviation obtained from five iterations. It is evident that DeepSAD alone outperforms SVDD for both metrics but is surpassed by the ADS method, resulting in a significant improvement in both correlation and RMSE scores. 
Indeed, employing ADS leads to a reduction in RMSE from 24.3 to 17.4.

The diversity regularisation improves performance when using both the norm of the embedding directly as HI and when employing the APAIC merging strategy on the 2DS embeddings. It does appear to provide enriched condition indicators that aid the APAIC procedure in finding more refined HIs.
The A2DS method results in a higher correlation score compared to ADS increasing from 0.973 to 0.977. It also reduces the RMSE from 17.4 to 14.04. 

Finally, it is important to highlight that in the real-time scenario, both RADS and RA2DS demonstrate similar overall scores when considering their respective standard deviations.

{\color{black}The A2DS and RA2DS approaches exhibit the best trendability performance compared to all methods, the third MK monotonicity after APAIC Feature and MOO, and the second-best prognosability performance, surpassed only by the MOO approache, which performs poorly  in terms of correlation and RMSE}. 

Figure~\ref{fig:y} presents the obtained HI for the three milling machines along with their respective ground truth HIs indicated by dotted lines. It demonstrates  that the DeepSAD models primarily emphasize  the regions associated with severe wear, while the APAIC merging methods reveal more complex trajectories. The high prognosability performance of MOO can be attributed to the fact that the HI values tend to remain constant when the milling machine is worn out. As shown in Table~\ref{Table:Milling}, the best fit is achieved with the RA2DS approache, where we observe both a strong correlation between the HI and a good alignment between the estimated HI values and the ground truth.}

{\color{black}Concerning the performance of our approach to the other semi-supervised learning approaches, we can see that for the case of the "pseudo-label" model \cite{nieves2022semi}, which is more oriented towards identifying the onset point where the anomaly starts, it is not able to capture the whole trajectory of the HI. This limitation is the reason for the poor performance observed here.  
Regarding the MOO approach \cite{moradi2023intelligent}, it yields high trendability and monotonicity values. However, the prognosability loss function differs from the prognosability score in Equation~\ref{prog}. As we lack access to ground truth labels during training, we utilize the standard prognosis loss function, where $t_P=T$ for all trajectories. This approach has a low RMSE and correlation scores since it is no longer possible to differentiate between the varying levels of wear at the end of the experiments.}

For more ablation studies showing the impact of $\beta$ and  the isotonic constraint in Equation~\ref{eq:time}, the impact of $\lambda$ in equation Equation~\ref{divDS} and the impact of the embedding size for ADS, please refer to \ref{appendix}.

\begin{table*}
\center
\begin{tabular}{c||ll||lll}
  & \multicolumn{2}{c||}{Criteria associated with ground truth} & \multicolumn{3}{c}{HI criteria without known ground truth}  \\
Method  & RMSE & Correlation & MK Monotonicity & Trendability & Prognosability \\
\hline
MO Feature& 47.12  & 0.776  & 0.854  & 0.561  & 0.462  \\
\textbf{APAIC Feature}& 24.29  & 0.928  & \textbf{0.994}  & 0.956  & 0.723  \\
SVDD \cite{tax2004support}& 31.40  & 0.793  & 0.934  & 0.875  & 0.550  \\
Pseudo-label \cite{nieves2022semi}& 24.37  & 0.927  & 0.958  & 0.878  & 0.563  \\
MOO \cite{moradi2023intelligent}& 39.54 $\pm$ 4.41& 0.898 $\pm$ 0.012& 0.990 $\pm$ 0.002& 0.970 $\pm$ 0.005& \textbf{0.923 $\pm$ 0.040}\\
DeepSAD& 24.24 $\pm$ 0.92& 0.890 $\pm$ 0.010& 0.949 $\pm$ 0.024& 0.976 $\pm$ 0.006& 0.576 $\pm$ 0.048\\
ADS& 17.42 $\pm$ 1.53& 0.965 $\pm$ 0.003& 0.967 $\pm$ 0.006& 0.976 $\pm$ 0.002& 0.789 $\pm$ 0.063\\
RADS& 17.41 $\pm$ 1.54& 0.965 $\pm$ 0.003& 0.966 $\pm$ 0.005& 0.976 $\pm$ 0.002& 0.789 $\pm$ 0.062\\
2DS& 22.59 $\pm$ 0.32& 0.904 $\pm$ 0.004& 0.976 $\pm$ 0.010& 0.976 $\pm$ 0.004& 0.688 $\pm$ 0.016\\
A2DS& \textbf{14.05} $\pm$ 0.74& \textbf{0.967} $\pm$ 0.006& 0.989 $\pm$ 0.002& \textbf{0.987} $\pm$ 0.003& 0.853 $\pm$ 0.011\\
RA2DS& \textbf{14.04} $\pm$ 0.73& \textbf{0.967} $\pm$ 0.006& 0.989 $\pm$ 0.002& \textbf{0.987} $\pm$ 0.003& 0.854 $\pm$ 0.011\\
\end{tabular} 
\caption{\color{black}Correlation, adapted RMSE 
 and standard HI criteria for the HI obtained using various methods. The proposed approaches are in bold case}\label{Table:Milling}
\end{table*}

\subsection{Comparison against supervised model}
The milling dataset, which includes ground truth HI, is widely regarded as an ideal dataset for supervised HI estimation. As a result, the majority of previous studies on this dataset utilized these labels to train various machine learning models\cite{zhou2022milling, gao2021new, wu2023cutting, liu2021tool, zhang2022dual}. In our work, we take a different approach by  not using the ground truth HI for training and instead  approximating labels. To provide a more relevant comparison, we focus on a recent study \cite{he2022milling}, which aligns with our work.  This study utilizes  data from all sensors and employs similar input features, specifically the Wavelet Packet Transform node output for each sensor modality.

{\color{black}Table~\ref{Table:super} presents the results obtained from various supervised methods as compared in \cite{he2022milling}. The correlation metric was not computed in the mentioned study. We focus on our best-performing approach, which is the RA2DS. Our semi-supervised approach demonstrates performance comparable to the top-performing supervised methods, only being surpassed by the stacked sparse AE proposed in \cite{he2022milling} which achieved an RMSE score of 12.7, while our approach yielded a score of 14.05.}
\begin{table}
\center
\begin{tabular}{c|cccc}
  & \multicolumn{4}{c}{RMSE} \\
Method  & c1 & c4 & c6 & Mean \\
\hline
MLP & 28.8 & 39.8 & 33.6 & 34.07  \\
CNN & 29.3 & 43.6 & 55.3 & 42.73  \\
LSTM  & 11.4 & 11.7 & 21.2 & 14.43  \\
RNN & 15.6 & 19.7 & 32.9 & 22.73  \\
BLSTM & 12.3 & 14.7 & 20.8 & 15.93  \\
Sparse Stacked AE & 9.3 & 14.0 & 14.8 & \textbf{12.70}  \\
RA2DS (semi-supervised) & 8.78  & 15.23 & 18.14 & \textbf{14.05} \\
\end{tabular} 
\caption{Comparison of our semi-supervised approach with supervised methods that use ground truth labels for training.}\label{Table:super}
\end{table}

\section{Application to spray coating monitoring}

In addition to the benchmark dataset evaluation, we also evaluate the proposed methodology on a real-world application: estimate the health condition of thermal spray for high quality surfaces coating. Thermal spray (TS) is an advanced technology used to efficiently coat surfaces and alter their mechanical and thermal properties. Some of the most common TS coating technologies include plasma spray, high-velocity oxy-fuel (HVOF), and Arc spray. Systems to which thermal spray coating is typically applied include jet engines, gas turbines, and automobile systems. Since these systems are safety-critical and mission-critical, coating quality is of significant importance. 

Amongst all TS technologies, Atmospheric Plasma Spray (APS) is the most versatile and widely used method capable of spraying a wide range of materials onto various  substrate materials. The spray gun is the most critical  component of the plasma spray coating system, and the nozzle-electrode pair is the most crucial part of the plasma spray gun. Currently, there is a lack of suitable tools that can monitor the quality and process stability of thermal spray coatings. The APS process is characterized by several parametric drifts and fluctuations occurring  at different time scales. These phenomena stem  from various  sources, including electrode wear and intrinsic plasma jet instabilities, as well as powder feeder fluctuations. In fact, the condition of the nozzle-electrode pair is important factor affecting the coating quality characteristics of plasma spray.

There has been limited research on the analysis of signals recorded during the coating process to monitor fluctuations in the thermal spray process. In a study conducted by \cite{kamnis2019aeroacoustics}, the HVOF process was investigated by collecting and analyzing airborne acoustic emissions (AAE) within the booth. This study relied on features extracted using FFT and was not tested with data from other guns of the same hardware or with different hardware and parameters. It was based on a very limited dataset from a single gun with specific hardware and parameters. Furthermore, it provided an estimation of the current process state but  could not be used for predictions. In another study by \cite{blair2015offline}, an offline  method was developed to determine the wear state of GH-type nozzles for Oelikon Metco 9MB plasma spray guns. This was achieved by recording and analyzing the acoustic signals generated by a controlled gas flow through each nozzle.

In this paper, we address the challenge of monitoring the health state and lifetime of the nozzle/electrode without requiring modifications to the spraying system (non-intrusive) and without disrupting the production plan (non-disruptive). Our approach involves   extracting relevant features from the High Frequency (HF) gun voltage signal during the coating process to assess the health state and lifetime of the gun nozzle and electrode. Typically, the lifetime of a nozzle ranges from 10 to 40 hours, depending on the stress imposed on the nozzle by the arc energy. The condition  of the nozzle significantly impacts coating quality. The objective is to estimate the nozzle's state in real-time and recommend  the optimal  time for nozzle replacement to achieve both the coating quality requirements and maximum hardware utilization.

\subsection{Dataset}

We collected high-frequency gun voltage for identifying nozzle wear. While plasma spray guns operate on direct current, the voltage fluctuates rapidly. These fluctuations and the mean voltage change with nozzle degradation. The condition monitoring system primarily employs a high frequency voltage sensor is used to record gun voltage generated by the thermal spray gun during operation. The data is stored, converted into a readable format, and then processed. The signal is  further connected to a high frequency DAQ (Data Acquisition System) for digital conversion. 

The sampling frequency was set to 50 kHz. We used an Oerlikon Metco F4 gun, with a 6mm nozzle, and recorded gun voltage for throughout the nozzle's lifetime for 4 nozzle-electrode pairs. Nozzle's lifetime varies depending on several controlled and uncontrolled conditions, such as  nozzle/electrode installation differences, gun setup, water temperature and more. In the experiments, we used 45/10.5 NLPM Ar/H2 with 650 A of current. We conducted incremental testing on the nozzles over a period of 15 to 45 hours of operation, assessing when the nozzle reached a “very worn” condition with the assistance of a domain expert. A significant  amount of hydrogen was employed to accelerate nozzle/electrode wear due to its constriction of the plasma core, resulting in higher energy density. We convert the high-frequency voltage signal into a Melspectrogram for feature extraction.

\subsection{Results}
Since we lack the ground truth, we propose employing alternative metrics in this section to evaluate the quality of our HI. The first metric measures the delay between the fault's onset time and when our estimated HI first reaches a value of 1, which we refer to as the "Delay." The second metric we employ is the Root Sum of Square Error (RSSE) calculated between the derived HIs and a HI constructed based on the real binary alarm profile. This constructed ground truth HI linearly increase from 0 to 1 until the fault's onset time, after which it remains constant at 1 until the end of the experiment. We cap the HI values at 1 for this specific metric to avoid penalizing HIs that could potentially exceed 1.
{\color{black}This score evaluates how closely the obtained HI aligns with a Remaining Useful Life (RUL) estimator, underscoring the potential of first extracting the HI and then using it for RUL prediction. The "Delay" score serves as a local metric, assessing our HI's ability to assess the end of life. Conversely, the RSSE score serves as a global metric, evaluating the overall quality of our HI's shape. We also take into account the monotonicity and prognosability criteria, but {\color{black} we excluded trendability due to the varying wear conditions among different nozzles and the significant variation in run-to-failure durations, which lead to different degradation trends.}}

{\color{black}We conducted a comparison among three methods: the DeepSAD model, the RADS, and the RA2DS. For RADS and RA2DS we consider a step size $\tau$ of 4 hours and we used time thresholds of $T_d$ and $T_f$ set to 5 hours and 3 hours, respectively.} We maintained consistent hyperparameters with the previous benchmark dataset and set $\lambda$ to a fixed value of 0.1. For both the DeepSAD and APAIC algorithms, we classified samples from the first 5 hours of machine operation as healthy. Conversely, we labeled samples recorded after the fault's onset time as abnormal.

Tables \ref{Table:oer1} and \ref{Table:oer2} present the Delay and RMSE scores for all four nozzles. 
DeepSAD failed to trigger alarms in nozzle two and three while RADS failed to trigger an alarm in nozzle four. RA2DS was able to detect the end of life for all four nozzles. On average, when an alarm was triggered, RA2DS exhibited the closest detection time to the fault onset, with an average delay of 1.4 hours. In contrast, RADS had an average alarm delay of 3.6 hours, and DeepSAD displayed an alarm delay of 7.8 hours. Regarding the RSSE scores, once again, RA2DS outperformed the other methods, with an average score of 12.7 and the DeepSAD model displayed the poorest performance with a score of 17.6. 

{\color{black}Table \ref{Table:oer3} presents the MK monotonicity and prognosability when the end-of-life period is determined based on the severe degradation onset time estimated by the domain expert. Although the prognosability score follows the trend of the previous two  scores, the DeepSAD score exhibits the highest level of monotonicity. In this case, it is primarily attributed to specific trajectories, such as the nozzle 4 for RADS, which does not demonstrate a monotonic trend. Additionally, the third nozzle of RA2DS displays two bumps at 10 and 14 hours, significantly impacting the final monotonicity index. }

\begin{figure}[htb!]
\centering
\begin{subfigure}[b]{0.4\textwidth}
        \centering \includegraphics[width=\textwidth]{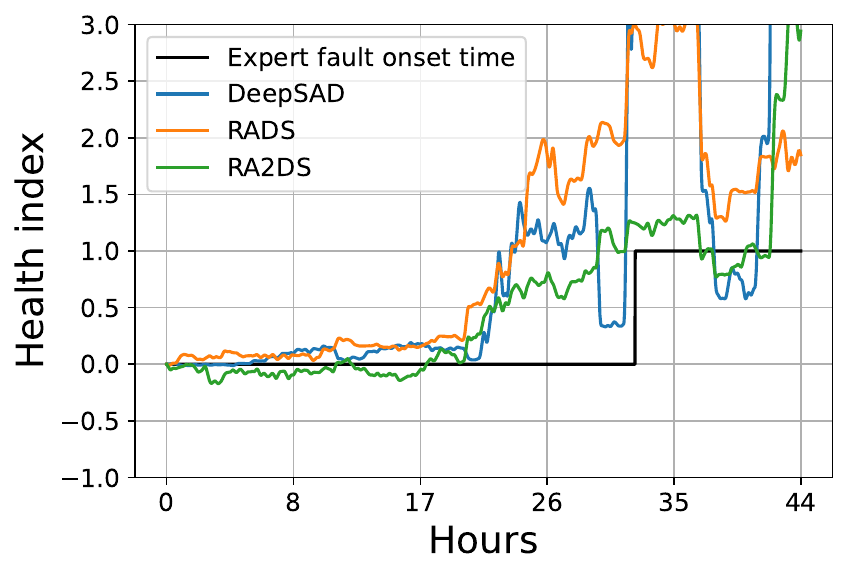}
        \caption{}
\end{subfigure}
\begin{subfigure}[b]{0.4\textwidth}
        \centering \includegraphics[width=\textwidth]{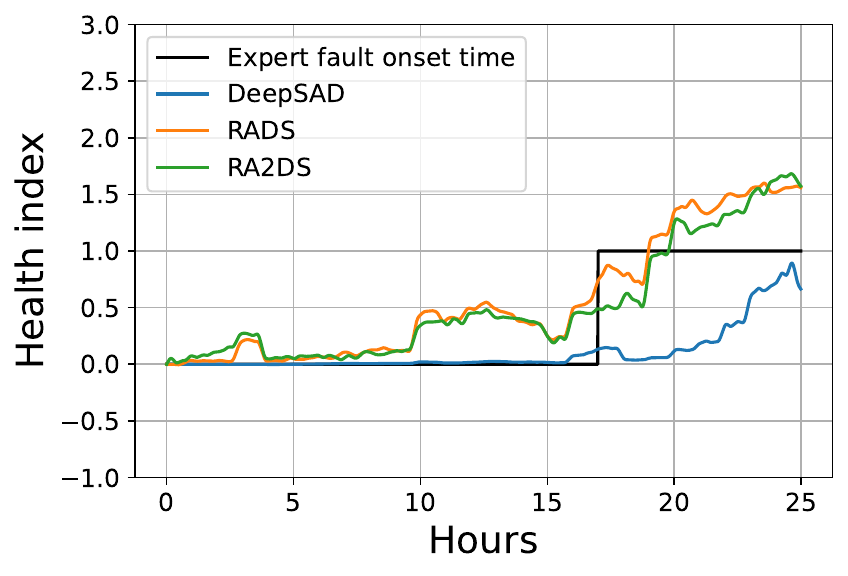}
        \caption{}
\end{subfigure}

\begin{subfigure}[b]{0.4\textwidth}
        \centering \includegraphics[width=\textwidth]{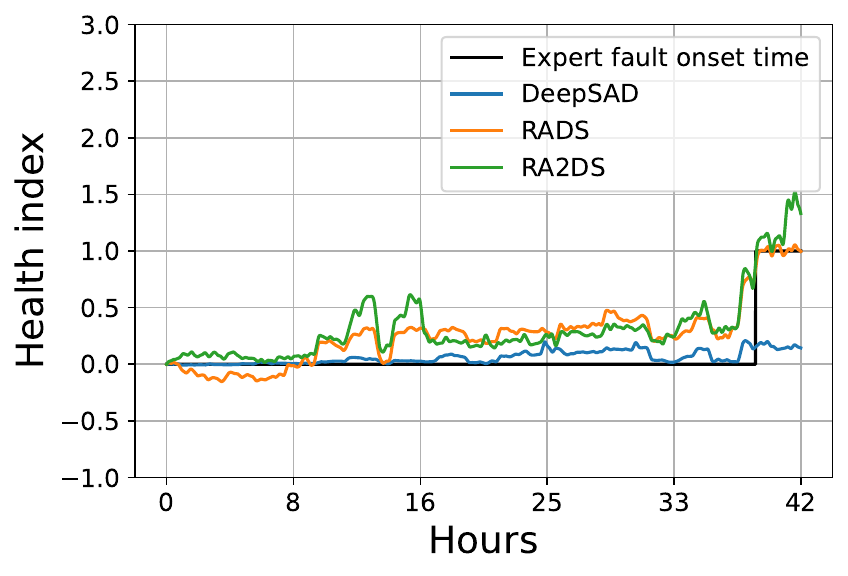}
        \caption{}
\end{subfigure}
\begin{subfigure}[b]{0.4\textwidth}
        \centering \includegraphics[width=\textwidth]{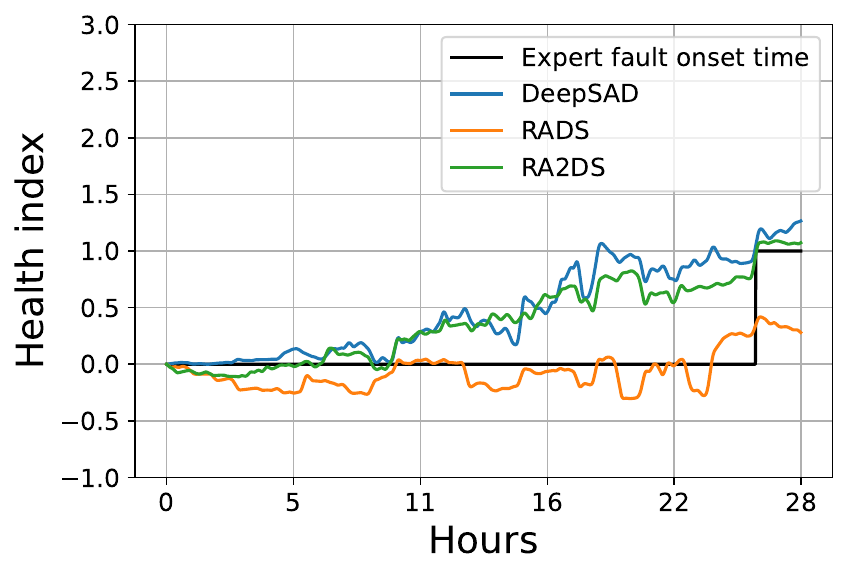}
        \caption{}
\end{subfigure}

    \caption{Comparison between estimated DeepSAD, ADS, A2DS HIS and binary alarm estimated by experts for the 4 nozzles lifecycles.  }\label{fig:y_oer}
\end{figure}

Figure \ref{fig:y_oer} shows the obtained HIs for each nozzle, with the expert binary labels represented in black. Among the four experiments, RA2DS stands out as the most reliable HI. It has a consistent evolution between 0 and 1. We cannot further assess the relevance of the obtained HI since we do not have access to the ground truth HI. Nevertheless, this score can be employed to evaluate the wear level, with the goal of estimating the RUL, triggering an alarm before the nozzle reaches a worn-out state, or extending the use of a nozzle if the associated HI remains low even after the typical replacement time.
\begin{table}
\center
\begin{tabular}{c|ccccc}
  & \multicolumn{5}{c}{Delay}  \\
Method  & N1 & N2 & N3 & N4  & \textbf{Mean}\\
\hline
DS &  8.62 & - & - & 6.948 & 7.784 \\
RADS &  8.55 & 2.02 & 0.33 & - & 3.632 \\
RA2DS &  2.50 & 2.83 & 0.10 & 0.086 & \textbf{1.379} \\
\end{tabular} 
\caption{
The "Delay" represents the time between the fault's estimated onset and the alarm raised by our estimated HI. A "-" indicates that no alarms were raised during that period.}\label{Table:oer1}
\end{table}

\begin{table}
\center
\begin{tabular}{c|ccccc}
  & \multicolumn{5}{c}{RSSE}  \\
Method  & N1 & N2 & N3 & N4  & \textbf{Mean}\\
\hline
DS &  13.74 & 23.80 & 26.86 & 5.962 & 17.588 \\
RADS &  9.05 & 10.39 & 16.23 & 26.204 & 15.466 \\
RA2DS &  15.03 & 11.88 & 16.40 & 7.588 & \textbf{12.724} \\
\end{tabular} 
\caption{The RSSE (Root Sum Square Error) is calculated between the synthetic ground truth and the estimated HI for each compared method. The HIs are adjusted to ensure they are capped at 1 during this analysis.}\label{Table:oer2}
\end{table}

\begin{table}
\center
\begin{tabular}{c|cccc||c|c}
  & \multicolumn{5}{c|}{MK Monotonicity}  &  Prog.  \\
Method  & N1 & N2 & N3 & N4  & \textbf{Mean}  & \\
\hline
DS &  0.90 & 0.96 & 0.80 & 0.96 & \textbf{0.91} & 0.36\\
RADS &  0.89 & 0.95 & 0.82 & 0.44 & 0.78 & 0.42 \\
RA2DS &  0.88 & 0.92 & 0.75 & 0.96 & 0.88 & \textbf{0.72}\\
\end{tabular} 
\caption{Averaged MK Monotonicity and Prognosability (Prog.).}\label{Table:oer3}
\end{table}

\begin{figure}[htb!]
\centering
\begin{subfigure}[b]{0.42\textwidth}
        \centering \includegraphics[width=\textwidth]{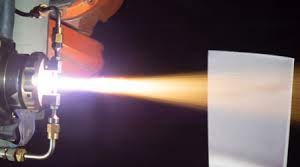}
        \caption{}
\end{subfigure}
\vspace{1cm}
\begin{subfigure}[b]{0.45\textwidth}
        \centering \includegraphics[width=\textwidth]{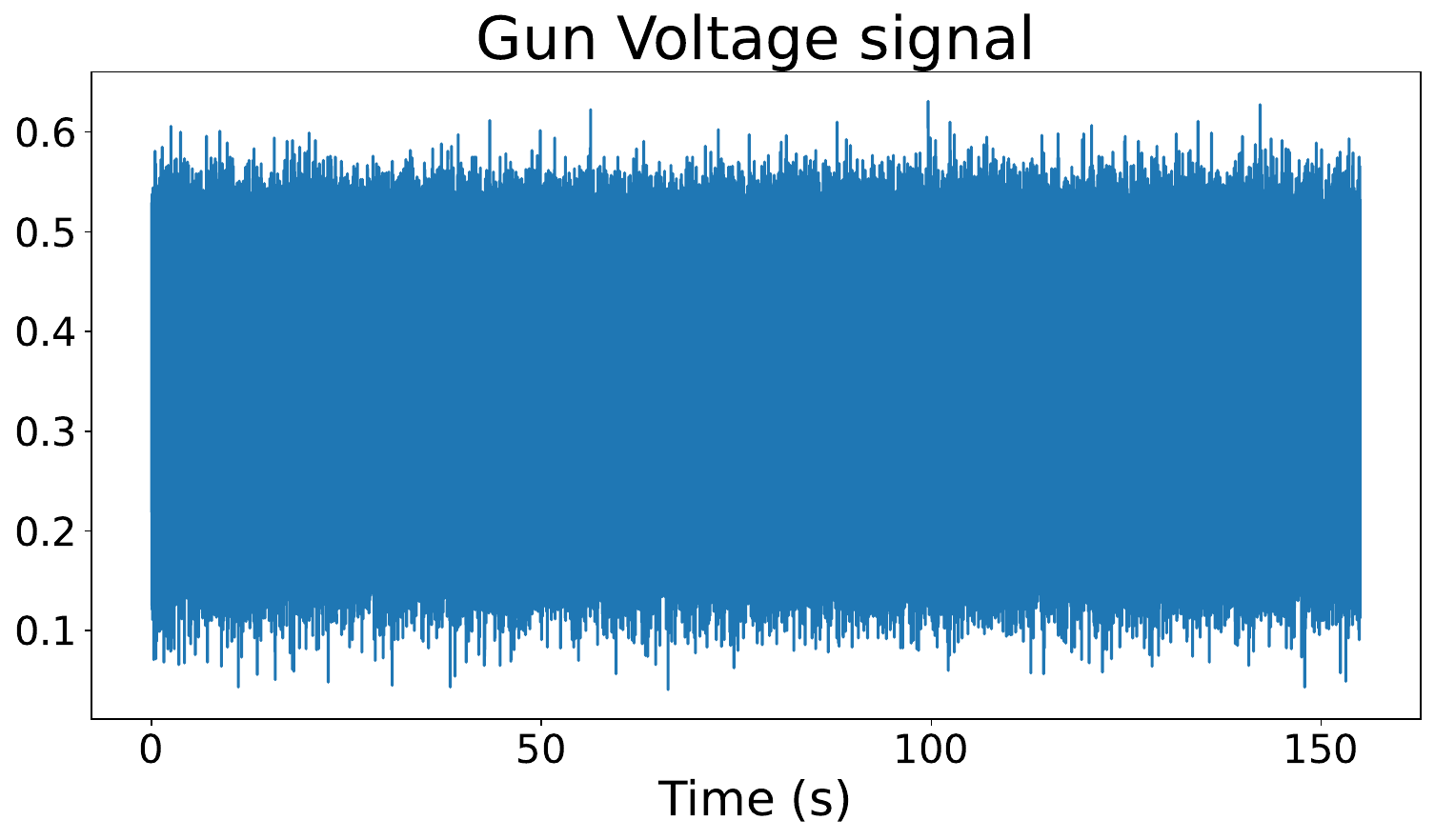}
        \caption{}
\end{subfigure}

\begin{subfigure}[b]{0.45\textwidth}
        \centering \includegraphics[width=\textwidth]{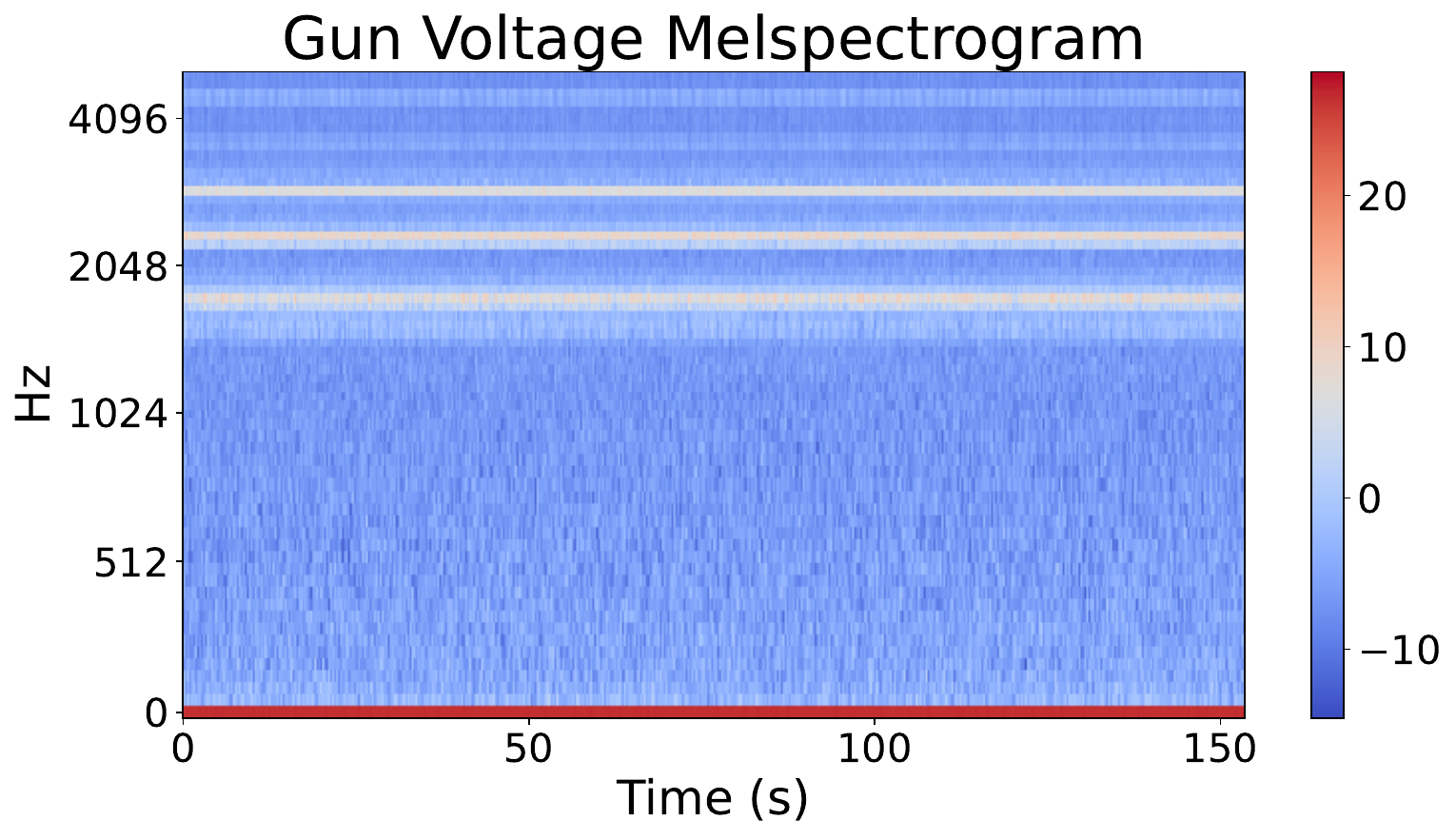}
        \caption{}
\end{subfigure}

\begin{subfigure}[b]{0.45\textwidth}
        \centering \includegraphics[width=\textwidth]{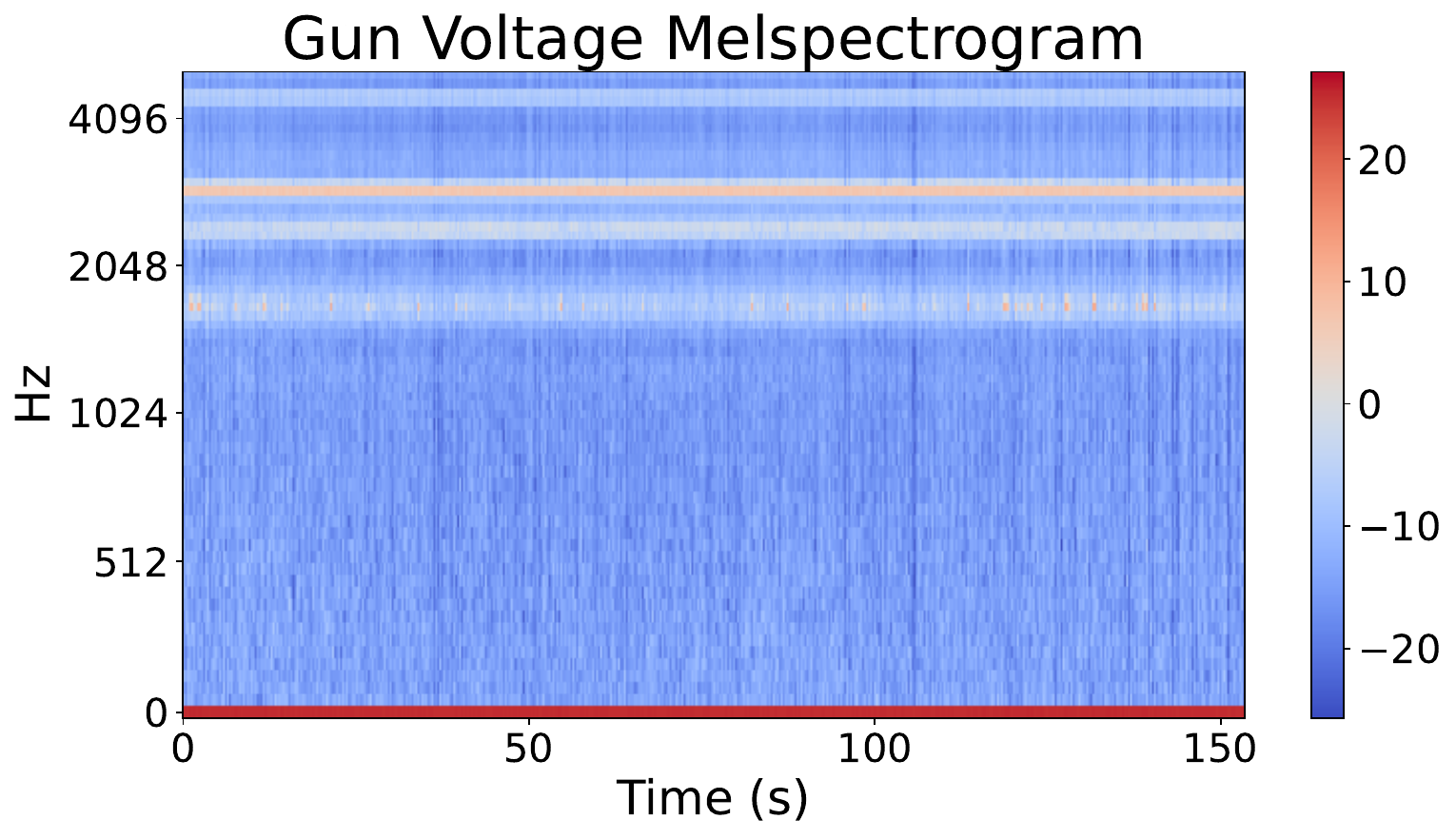}
        \caption{}
\end{subfigure}
    \caption{(a) Spray coating system - (b) High frequency volage signal - (c)-(d) Melspectrogram of the voltage signal repsectively at the begin and the end of the lifecycle.}\label{fig:oer}
\end{figure}

\section{Conclusion}

In this study, we introduced a HI construction method based on a semi-supervised anomaly detection approach called DeepSAD. {\color{black}Unlike the fully supervised approach, which requires a measurement of the machine's actual health state, a process that is often not feasible in practical applications.} Often, it is only feasible to acquire labels for the beginning or end of a system's lifecycle through available data. Indeed, there are healths states that are easier to assess: when the system is new and can be assumed to be healthy, or when the system fail and we are sure it is degraded. Thus, we propose a semi-supervised approach for HI construction. Our approach involves enhancing the DeepSAD embedding to generate condition indicators associated with various wearing within the system. These indicators are then integrated to create the HI using a novel alternating projection algorithm that ensures a normalized and monotonically increasing HI.

We evaluated the robustness of our approach using the PHME 2010 milling dataset, a benchmark dataset with ground truth HI values. Our findings demonstrate that our approach not only produces HIs that correlate with ground truth data but also ensures that the estimated HI values correspond to the relative wear states of different machines. Furthermore, we evaluate the applicability our approach to a real-world application, monitoring the wear states of thermal spray coatings with high-frequency voltage sensors. Our results indicate that our method yields HIs that consistently increase when expert detect that the system approaches the end of life.

Potential future directions for this research include exploring the application of the APAIC algorithm for feature merging in scenarios involving high-dimensional features or data from different modalities. Another avenue of investigation involves combining the APAIC and DeepSAD models into an end-to-end learning approach for the direct estimation of a robust HI.

\section*{Aknowledgments}
This study was financed by the Swiss Innovation Agency (lnnosuisse) under grant number: 47231.1 IP-ENG.

We would like to express our gratitude to Ehsan Fallahi for carrying out the data collection for the spray coating application, and Ron Molz for his expert opinion on data analysis and labeling the data.

\bibliographystyle{elsarticle-num}
\bibliography{bib}

\appendix
\section{Ablation study }\label{appendix}
This section relate to the same experiments as in Section~\ref{MillingExp}. We study the influence of the different hyper-parameters from the ADS and A2DS methodology.
\subsection{Impact of the parameters of APAIC }
We explore the ADS methodology for different values of $ \beta $ in Equation~\ref{eq:time} using the Ridge regularization. We also study the presence or absence of the Isotonic constraint. The results are presented in Table~\ref{App:A1}. We can see that without both the Ridge regularisation ($\beta=0$) and the Isotonic constraint the algorithm does not succeed to converge. Overall the results are stable for different values of $\beta$ with exactly the same correlation score and fairly similar RMSE score. 

\subsection{Impact of the size of the embedding $K$ in DeepSAD }
We investigate the embedding size of DeepSAD, denoted as $\mathbf{Y} \in \mathbb{R}^{F \times K}$, for various values of $K$. The results are presented in Table~\ref{App:A2}. Once more, the results remain stable regardless of the value of this hyperparameter.

\subsection{Impact of the diversity parameters $\lambda$ in 2DS }
We examine the impact of the diversity regularization parameter $\lambda$ as defined in Equation~\ref{divDS}. The outcomes are displayed in Table~\ref{App:A3}. For values of $\lambda$ greater than or equal to 0.01, we select the results with the lowest loss after five different initializations, as the algorithm yields varied results depending on the initialization. We defer the investigation of this issue for future research. It seems that the value of $\lambda$ needs to be carefully balanced. When it becomes too high, the parameters of the DeepSAD model become negligible in comparison to the diversity loss, which results in trajectories that cannot be considered as reliable condition indicators. The value $\lambda = 0.001$ corresponds to a balancing parameter that aligns the magnitudes of the DeepSAD model loss and diversity loss for this experiment."

\begin{figure*}[htb!]
\centering
\begin{subfigure}[b]{0.35\textwidth}
        \centering \includegraphics[width=\textwidth]{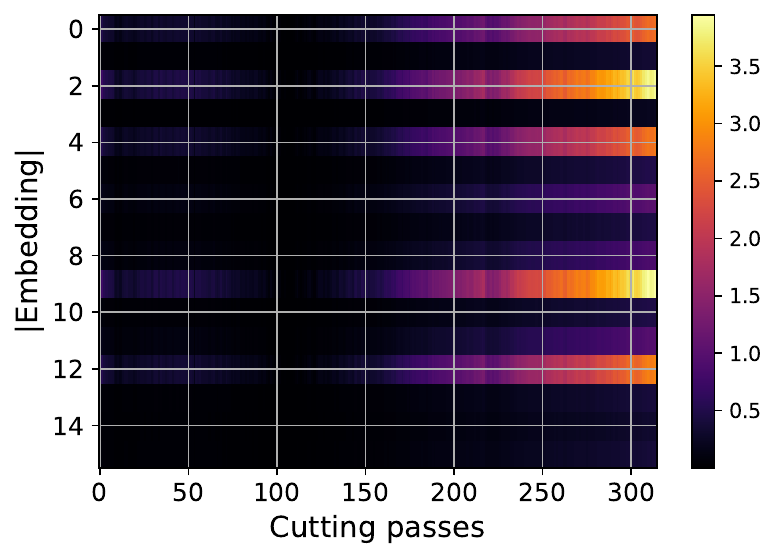}
        \caption{}
\end{subfigure}
\begin{subfigure}[b]{0.35\textwidth}
        \centering \includegraphics[width=\textwidth]{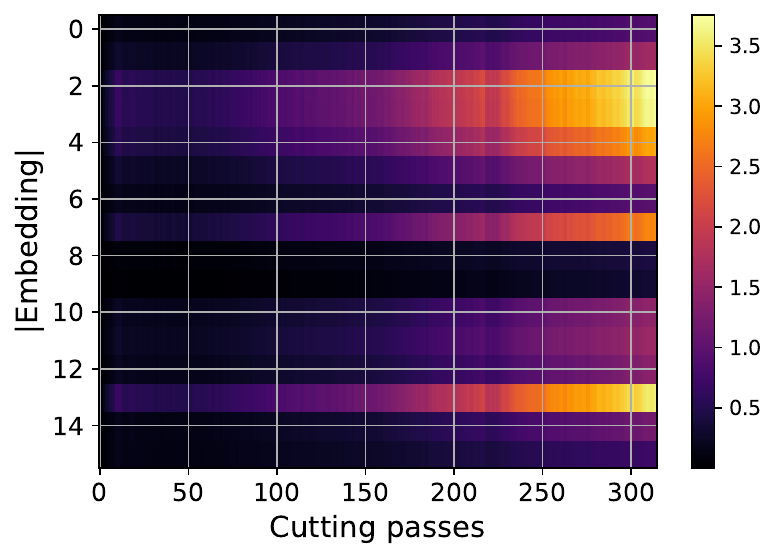}
        \caption{}
\end{subfigure}

\begin{subfigure}[b]{0.35\textwidth}
        \centering \includegraphics[width=\textwidth]{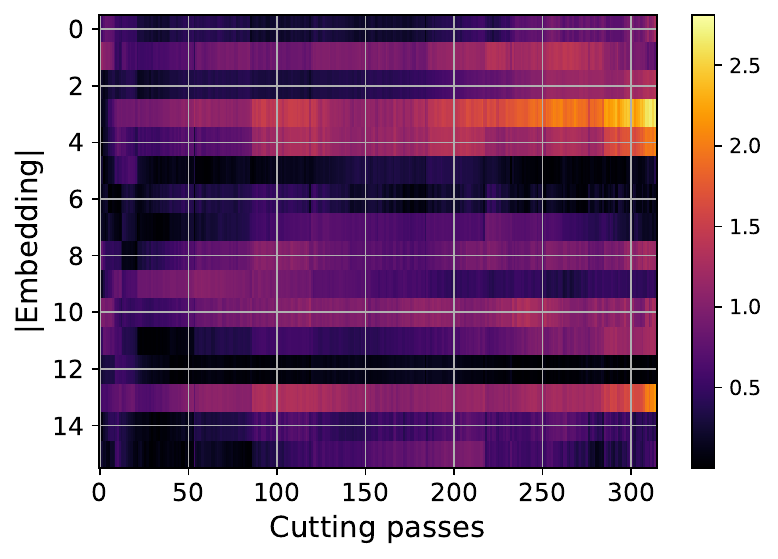}
        \caption{}
\end{subfigure}
\begin{subfigure}[b]{0.35\textwidth}
        \centering \includegraphics[width=\textwidth]{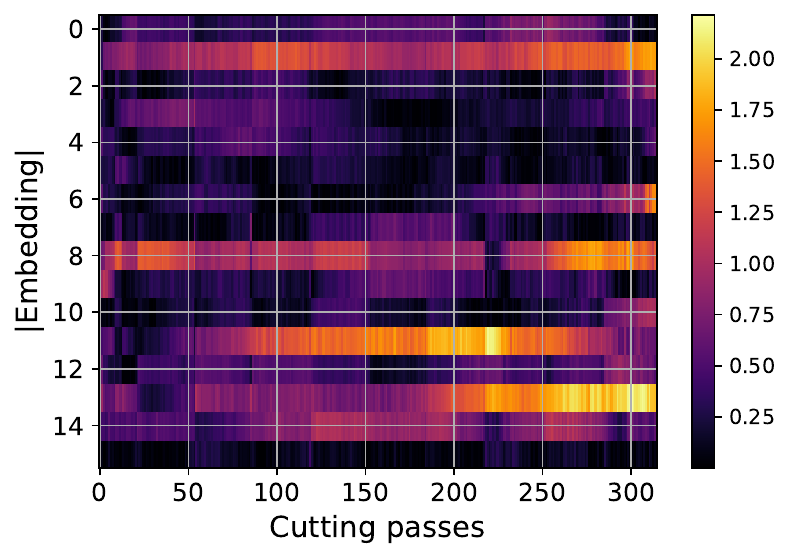}
        \caption{}
\end{subfigure}
    \caption{Absolute value of the lifecycle embedding obtained for the test c1 dataset for different diversity regularisation(a) 0  (b) 0.001 (c) 0.01 (d) 1 }\label{App:A4}
\end{figure*}

In Figure~\ref{App:A4}, we present the absolute embeddings acquired from the test dataset c6 for four distinct values of $\lambda$. In the case of $\lambda=0$, most trajectories display precisely the same pattern. As we introduce $\lambda=0.001$, some condition indicators activate at different times, yielding more diverse patterns. When $\lambda=1$, the embedding exhibits varying activation periods, effectively segmenting the time axis into different clusters. Although these diverse trajectories hold potential for future investigations, they appear noisier and more challenging to integrate for the APAIC algorithm.

\begin{table*}
\center
\begin{tabular}{cc|cccc|cccc}
$\beta$ &  Isotonicity  & \multicolumn{4}{c|}{RMSE} & \multicolumn{4}{c}{Correlation} \\
& & c1 & c4 & c6 & \textbf{Mean} & c1 & c4 & c6 & \textbf{Mean}\\
\hline
 $ \beta=0 $ & no & 147.33 & 96.25 & 145.28 & 129.62 & -0.514 & -0.791 & -0.639 & -0.648 \\
 $ \beta=0 $ & yes  & 26.73 & 30.97 & 16.58 & 24.76 & 0.979 & 0.897 & 0.962 & 0.946 \\
 $ \beta=0.001 $ & yes & 7.77 & 30.40 & 17.23 & 18.47 & 0.969 & 0.971 & 0.977 & 0.972 \\
 $ \beta=0.01 $ & yes & 7.41 & 29.88 & 17.49 & 18.26 & 0.969 & 0.971 & 0.977 & 0.972 \\
 $ \beta=0.1 $ & yes & 7.83 & 28.28 & 18.66 & 18.26 & 0.969 & 0.971 & 0.977 & 0.972 \\
 $ \beta=1 $ & no & 8.92 & 27.88 & 18.99 & 18.60 & 0.969 & 0.971 & 0.977 & 0.972 \\
 $ \beta=1 $ & yes & 8.06 & 27.80 & 18.87 & 18.24 & 0.969 & 0.971 & 0.977 & 0.972 \\
\end{tabular} 
\caption{Impact of the Isotonic constraint and Ridge regularisation hyperparameter $\beta$ applied on one realisation of the ADS algorithm}\label{App:A1}
\end{table*}

\begin{table*}
\center
\begin{tabular}{c|cccc|cccc}
  & \multicolumn{4}{c|}{RMSE} & \multicolumn{4}{c}{Correlation} \\
Dimension  & c1 & c4 & c6 & \textbf{Mean} & c1 & c4 & c6 & \textbf{Mean}\\
\hline
8& 7.76 & 19.03 & 12.26 & 13.01 & 0.968 & 0.965 & 0.980& 0.971 \\
16& 8.78 & 15.23 &  18.14 & 14.05 & 0.953 & 0.972 & 0.976 & 0.967 \\
32& 10.58 & 20.20 & 10.69& 13.83 & 0.966 5& 0.965 & 0.981 & 0.971\\
64& 9.91 & 19.38 & 9.87 & 13.06 & 0.963 & 0.961 2& 0.980 & 0.968 \\
\end{tabular} 
\caption{Impact of the dimension of the ADS embedding}\label{App:A2}
\end{table*}

\begin{table*}
\center
\begin{tabular}{c|cccc|cccc}
  & \multicolumn{4}{c|}{RMSE} & \multicolumn{4}{c}{Correlation} \\
$\lambda$  & c1 & c4 & c6 & \textbf{Mean} & c1 & c4 & c6 & \textbf{Mean}\\
\hline
$\lambda=$0 & 9.78 & 23.25 & 19.22 & 17.42 & 0.960 & 0.968 & 0.967 & 0.965 \\
$\lambda=$0.001& 8.78 & 15.23 &  18.14 & 14.05 & 0.953 & 0.972 & 0.976 & 0.967 \\
$\lambda=$0.01& 12.46 & 28.90 & 17.82 & 19.73 & 0.971 & 0.948 & 0.975 & 0.965 \\
$\lambda=$1& 18.80& 28.53 & 17.70 & 21.68 & 0.968 & 0.931 & 0.958 & 0.952 \\
\end{tabular} 
\caption{Impact of the dimension of the Diversity regularisation}\label{App:A3}
\end{table*}

\end{document}